\documentclass[11pt,a4paper]{article}

\usepackage{amssymb}
\usepackage[table]{xcolor}
\usepackage{amsmath}
\usepackage{booktabs}
\usepackage{listings}
\usepackage{tabularx}
\usepackage{hyperref}
\usepackage{graphicx}
\usepackage{wrapfig}
\usepackage[capitalise,nameinlink,noabbrev]{cleveref}

\usepackage{times,latexsym}
\usepackage{url}
\usepackage[T1]{fontenc}
\usepackage{cleveref}
\usepackage{graphics}
\usepackage{nicefrac}
\usepackage{enumitem}

\usepackage[acceptedWithA]{tacl2021v1}

\usepackage{tacl2021v1} %

\usepackage{xspace,mfirstuc,tabulary}

\newif\iftaclinstructions
\taclinstructionsfalse %
\iftaclinstructions
\renewcommand{\confidential}{}
\renewcommand{\anonsubtext}{(No author info supplied here, for consistency with
TACL-submission anonymization requirements)}
\newcommand{\instr}
\fi

\iftaclpubformat %

\else

\fi

\usepackage{amsmath,amsfonts,bm}

\def\eqref#1{equation~\ref{#1}}

\def\1{\bm{1}}

\DeclareMathAlphabet{\mathsfit}{\encodingdefault}{\sfdefault}{m}{sl}
\SetMathAlphabet{\mathsfit}{bold}{\encodingdefault}{\sfdefault}{bx}{n}

\newcommand{\todospacer}{\ifhmode\space\fi}

\usepackage{xspace}
\newcommand{\eg}{\textit{e.g.}\xspace}
\newcommand{\ie}{\textit{i.e.}\xspace}

\newcommand{\wrt}{wrt\xspace}
\usepackage{subcaption}

\newcolumntype{Y}{>{\raggedright\arraybackslash}X}
\definecolor{lightgray}{gray}{0.96}

\crefname{figure}{Fig.}{Figs.}
\crefname{table}{Tab.}{Tabs.}
\crefname{equation}{Eq.}{Eqs.}
\crefname{section}{Sec.}{Secs.}
\crefname{subsection}{Sec.}{Secs.}
\crefname{subsubsection}{Sec.}{Secs.}
\crefname{appendix}{App.}{Apps.}
\crefname{algorithm}{Alg.}{Algs.}
\crefname{theorem}{Thm.}{Thms.}
\crefname{lemma}{Lem.}{Lems.}
\crefname{proposition}{Prop.}{Props.}
\crefname{corollary}{Cor.}{Cors.}
\crefname{definition}{Def.}{Defs.}

\definecolor{name1}{HTML}{B22222} %
\definecolor{name7}{HTML}{8B4513} %
\definecolor{name3}{HTML}{006400} %
\definecolor{name4}{HTML}{556B2F} %
\definecolor{name5}{HTML}{A0522D} %
\definecolor{name6}{HTML}{B8860B} %
\definecolor{name2}{HTML}{800080} %
\definecolor{name8}{HTML}{6A3D9A} %
\definecolor{name9}{HTML}{008080} %
\definecolor{name10}{HTML}{1F4E79} %

\title{Teaching Language Models to Faithfully Express their Uncertainty}

\author{
  Bryan Eikema\textsuperscript{*} \\
  University of Amsterdam \\
  \texttt{b.eikema@uva.nl}
  \And
  Evgenia Ilia\textsuperscript{*} \\
  University of Amsterdam \\
  \texttt{e.ilia@uva.nl}
  \And
  José G. C. de Souza \\
  Outsystems \\
  \texttt{jose.souza@outsystems.com}
  \AND
  Chrysoula Zerva \\
  Instituto de Telecomunicações\\
  Instituto Superior T\'ecnico, Universidade de Lisboa\\
  \texttt{chrysoula.zerva@tecnico.ulisboa.pt}
  \And
  Wilker Aziz \\
  University of Amsterdam \\
  \texttt{w.aziz@uva.nl}
}

\author{%
  Bryan Eikema\textsuperscript{*,\,$\diamond$} \quad
  Evgenia Ilia\textsuperscript{*,\,$\diamond$} \quad
  José G. C. de Souza\textsuperscript{$\triangle$} \quad
  Chrysoula Zerva\textsuperscript{$\ltimes$} \quad
  Wilker Aziz\textsuperscript{$\diamond$}
  \\
  \textsuperscript{$\diamond$}University of Amsterdam \quad
  \textsuperscript{$\triangle$}Outsystems \quad
  \textsuperscript{$\ltimes$}Instituto de Telecomunicações
  \\
  \texttt{\href{mailto:b.eikema@uva.nl}{b.eikema@uva.nl}} \quad
  \texttt{\href{mailto:e.ilia@uva.nl}{e.ilia@uva.nl}} \quad
    \texttt{\href{mailto:jose.souza@outsystems.com}{jose.souza@outsystems.com}}
  \\
  \texttt{\href{mailto:chrysoula.zerva@tecnico.ulisboa.pt}{chrysoula.zerva@tecnico.ulisboa.pt}}  \quad
  \texttt{\href{mailto:w.aziz@uva.nl}{w.aziz@uva.nl}} 
}

\date{}

\begin{document}

\maketitle

\begingroup
\renewcommand\thefootnote{}\footnote{\textsuperscript{*} Equal contribution.}%
\addtocounter{footnote}{-1}%
\endgroup

\begin{abstract}

Large language models (LLMs) often miscommunicate their uncertainty: repeated queries can produce divergent answers, yet generated responses are typically unhedged or hedged in ways that do not reflect this variability.
This conveys unfaithful information about the uncertain state of the LLMs' knowledge,  creating a \emph{faithfulness gap} that affects even strong LLMs. %
We introduce Faithful Uncertainty Tuning (FUT): a fine-tuning approach that teaches instruction-tuned LLMs to express uncertainty faithfully without altering their underlying answer distribution. 
We construct training data by augmenting model samples with \emph{uncertainty hedges} (\textit{i.e.} verbal cues such as `possibly' or `likely') aligned with sample consistency, requiring no supervision beyond the model and a set of prompts.
We evaluate FUT on open-domain question answering (QA) across multiple models and datasets. %
Our results show that FUT substantially reduces the faithfulness gap, %
while preserving QA accuracy and introducing minimal semantic distribution shift.
Further analyses demonstrate robustness across decoding strategies, choice of hedgers, and other forms of uncertainty expression (\emph{i.e.} numerical).
These findings establish FUT as a simple and effective way to teach LLMs to communicate uncertainty faithfully.

\end{abstract}

\begin{figure}[t!]
    \centering
\includegraphics[width=1\linewidth]{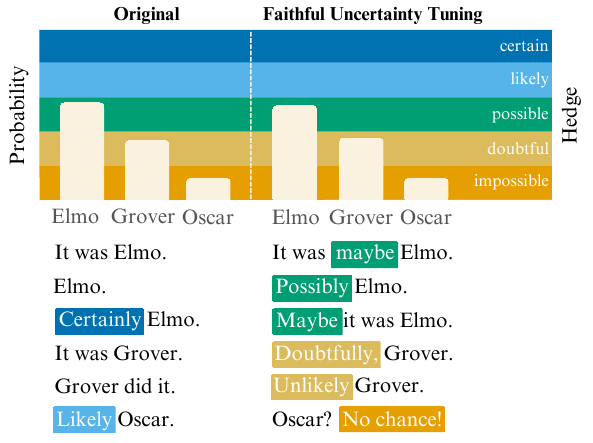}

\caption{
We want to know \emph{Who pushed Big Bird?}
and, for transparency’s sake, we expect a good response to be suggestive of uncertainty in the model’s state of knowledge. Each plot shows the beliefs of a model expressed as probabilities over clustered responses, with sampled responses shown below the plot. We segment the vertical axes in 5 intervals, and highlight hedge phrases that humans regard as coherent with those intervals of probabilistic belief. The original model (left) generates responses that are either unhedged or hedged unfaithfully (\textit{e.g.} the model’s belief in Elmo is in the ballpark of `possible’, but responses blaming Elmo suggest `certainty’), leading to misinformation. Our method (right) adapts
the model so that responses are faithfully hedged while closely preserving the original beliefs.
}
    \label{fig:sketch2}
\end{figure}

\section{Introduction}
Large language models (LLMs) can generate responses that, alongside providing an answer to a query, are hedged to convey their `confidence' in the answer \citep{linteaching,kadavath2022language}. 
A recent notion of confidence \citep{kuhn2023semantic,manakul-etal-2023-selfcheckgpt,tian2024finetuning} quantifies the semantic consistency of a response with respect to other likely responses under the model distribution.
Hedging grounded in this notion conveys information that is \emph{faithful} to the model's own beliefs, offering a window into what the model `knows' \citep{farquhar2024detecting, Joo2025BlackBoxHD}.
Hedging can help users gauge their trust in generated answers and adapt their interactions accordingly %
\citep{wang2025impact, kim2024m, steyvers2025large}.
Yet producing faithful hedges remains difficult: \citet{yona-etal-2024-large} show that LLMs exhibit a \emph{faithfulness gap}, often sounding confident despite substantial internal disagreement, and that even modest improvements require heavy prompt engineering \citep{liu2025metafaith}.
This raises a key question: 
can we adapt a model such that its responses are hedged to convey the model's beliefs, while keeping those beliefs unchanged? 
We illustrate these desiderata in \cref{fig:sketch2}.

To address this, we introduce Faithful Uncertainty Tuning (FUT), a fine-tuning method that teaches instruction-tuned LLMs to express uncertainty faithfully while preserving the model's underlying beliefs.
We achieve this by constructing a fine-tuning dataset from unbiased model samples, augmented with appropriate hedges derived from sample consistency, requiring no external supervision beyond access to a set of input prompts and the model itself.
This separates two concerns: \emph{correctness}, learned during initial model training and governs whether the distribution captures factual answers, and \emph{uncertainty communication}, which FUT addresses by aligning hedges with that distribution.
Essentially, FUT equips models with the ability to communicate \emph{instance-level} confidence, \ie confidence in each individual prediction.

We study FUT in the setting of open-domain question answering (QA), applying it to multiple instruction-tuned LLMs. 
We experiment with two strategies for introducing hedges: interweaving them naturally into the answer, or postfixing them while leaving the original response intact. 
We compare FUT against the base models both under typical prompting and when prompted to explicitly elicit uncertainty hedging. 
Our evaluation focuses on three aspects: faithfulness, measured by conditional mean faithful generation \citep[cMFG;][]{yona-etal-2024-large}, and QA accuracy, both evaluated across multiple QA datasets, as well as semantic distribution shift, quantified as the total variation distance between the semantic distributions of answers from base and fine-tuned models.

We find FUT to substantially improve faithfulness 
across QA datasets and models, while preserving task performance and introducing minimal semantic distribution shift.
We further analyse
\textit{i)} linguistic vs. numerical uncertainty expressions,
\textit{ii)} remaining failure modes, 
\textit{iii)} the choice of hedge phrases,
and \textit{iv)} faithfulness of various decoding strategies.
Our contributions are as follows: we introduce Faithful Uncertainty Tuning as, to our knowledge, the first fine-tuning approach that enables instruction-tuned LLMs to communicate instance-level confidence \emph{faithfully} without altering their underlying distribution; we demonstrate substantial gains in faithfulness alongside preserved task performance; and we provide extensive analyses that demonstrate the robustness and generality of our fine-tuning method, which requires only the model itself and modest computational effort.
Taken together, these results establish FUT as a simple and effective approach for teaching LLMs to communicate their uncertainty faithfully to their own beliefs.\looseness-1 \footnote{We release experimental code at \url{https://github.com/probabll/uncertainty-tuning}.}

\section{Measuring Faithfulness}
\label{sec:background}

A model's response $y$ to a prompt $x$ can be regarded as a collection $\mathcal A(y)$ of one or more assertions whose linguistically expressed confidence---henceforth referred to as `decisiveness'---may or may not convey accurate information about the uncertain state of knowledge in which $y$ was generated. 
For each assertion $a \in \mathcal A(y)$, \citet{yona-etal-2024-large} quantify the gap between a model-based notion of \emph{confidence} $C(a)$ in the assertion and the \emph{decisiveness} $D(a)$ with which %
the assertion is expressed.
They regard a response $y$ as \emph{faithfully hedged to express the model's own uncertainty} when, on average across assertions, this gap is small. 
This is to say that they take
\begin{equation}
    F(y|x) = 1 - \frac{1}{|\mathcal{A}(y)|} \sum_{a \in \mathcal{A}(y)} |D(a) - C(a)|
\end{equation}
as the \textbf{faithful generation score} of $y$ given $x$ \wrt the model and its probabilistic beliefs.
We largely adopt their formulation, as explained next.

Following \citet{yona-etal-2024-large}, we identify the model's \textbf{confidence} in an assertion $a$ with the rate at which responses sampled from the model are expected not to contradict $a$:\footnote{\textbf{Notation.} We use $Y$ to denote a random variable taking on outcomes in the space $\mathcal Y$ of all possible responses. The distribution of $Y$ is given by an LLM, conditioned on a prompt $x$. The expected value under the model of some function $\phi(y)$ of responses is denoted $\mathbb E[\phi(Y)|x]$ and defined as $\sum_{y \in \mathcal Y} \phi(y)p(y|x)$, where $p(y|x)$ is the probability which the model assigns to a candidate response $y$ given $x$. The binary operator $\bot$ stands for logical contradiction.} 
\begin{equation}
    C(a) = 1 - \mathbb{E}\left[a \bot Y | x\right] ~,
    \label{eq:faithful-confidence}
\end{equation}
with the contradiction rate  estimated via Monte Carlo (MC; \ie we draw independent samples $y_1, \ldots, y_N$ from the model and compute a sample mean $\nicefrac{1}{N}\sum_{n=1}^N[a \bot y_n]$). 
To assess assertions and responses for contradictions, we follow the original work and use a natural language inference (NLI) judge (implementation in \cref{app:judge-prompts}).

\textbf{Decisiveness} captures how certain an assertion `sounds' to a reader, who forms their judgment based on the speaker's use of so-called \emph{uncertainty hedges} \citep[\ie, verbal cues such as `possibly' and `for sure';][]{mielke2022reducing}.
We work with correspondences between uncertainty hedges and intervals of probabilistic belief uncovered by studies in psycholinguistics \citep{vogel2022interpretation}; specifically, we use the mapping shown in \cref{tab:conf_to_hedger}. 
As in \citet{yona-etal-2024-large}, we few-shot prompt an LLM to jointly extract assertions and their decisiveness levels. The prompt includes the input $x$, a response $y$, and a few examples that demonstrate the task (implementation in \cref{app:judge-prompts}).

For corpus-level evaluation, \citet{yona-etal-2024-large} introduce the \textbf{conditional mean faithful generation (cMFG) score}.  
First, prompt-response pairs are grouped into bins based on the average confidence of the response's assertions (\ie, $(x,y)$ is binned based on $\nicefrac{1}{|\mathcal A(y)| \sum_{a \in \mathcal A(y)}C(a)}$), covering the intervals $[0,0.1], (0.1,0.2], \ldots (0.9, 1]$. 
The cMFG score is then the macro-average of the faithful generation scores across these bins. %
The score lies in $[0,1]$: $1$ indicates perfect faithfulness (\ie, decisiveness exactly matches confidence), $0.5$ corresponds to the worst-case (\ie, decisiveness is completely uncorrelated with confidence, such as when  all assertions are hedged decisively or at random), and $0$ reflects maximal misalignment (\ie, decisiveness is  inverted \wrt  confidence). %

We rely and build upon this framework in two ways. First, in \cref{sec:faithful-uncertainty-tuning} we construct a fine-tuning dataset of faithfully hedged responses, which we then use to fine-tune a model to natively produce faithful responses. Second, in \cref{sec:experiments} we adopt cMFG as the evaluation framework of choice to assess the faithfulness gap of our trained models.

\section{Faithful Uncertainty Tuning (FUT)}
\label{sec:faithful-uncertainty-tuning}

We introduce Faithful Uncertainty Tuning (FUT), a fine-tuning approach that teaches a model to express faithful uncertainty hedges without altering its underlying distribution over asserted content (\cref{fig:sketch2}).
The essence of FUT is a careful data generating process that creates a fine-tuning dataset exhibiting all the necessary features for models to learn faithful hedging while preserving  beliefs over asserted content.

\paragraph{Faithful response generation.} Given a prompt $x$, we generate a faithfully hedged response \emph{by construction} in 5 steps: (1) draw a candidate response $y$ given $x$; (2) analyse this response into a collection of assertions $\mathcal{A}(y)$; (3) for each assertion $a \in \mathcal{A}(y)$, estimate confidence $C(a)$ as explained in \cref{sec:faithful-uncertainty-tuning} (\cref{eq:faithful-confidence}); (4) hedge each assertion $a$ to convey decisiveness coherently with $C(a)$; and (5) rewrite the hedged assertions into a new response (we detail steps 4--5 in the next paragraph).  We repeat all five steps to produce faithful responses for an entire dataset of prompts. 
In fact, we repeat this entire procedure $S$ times for each and every prompt $x$, in order to better represent the model's beliefs over asserted contents given $x$. To spare resources, we reuse the $S$ candidate responses for any one $x$ in the estimation of contradiction rates (that is, as sampled realisations of $Y$ in \cref{eq:faithful-confidence}), but we perform leave-one-out estimation to avoid bias.
For that same reason, namely, to stay consistent with the model's `native' probabilistic beliefs, we employ unbiased sampling, as opposed to a decoding algorithm that injects biases of its own (\eg, greedy decoding), in step (1) and, by extension, in step (3). It is because our hedges in (4) are derived directly from an unbiased estimate of confidence in (3), that this procedure yields faithful responses by construction.

\begin{table}[t]
\centering
\scriptsize
\renewcommand{\arraystretch}{1.2}
\setlength{\tabcolsep}{6pt} 
\begin{tabular}{l l l}
\toprule
\textbf{Confidence} & \textbf{Uncertainty Hedge} & \textbf{Postfix Phrase} \\
\midrule
$[0.0, 0.1]$     & Impossible         & It is impossible. \\
$(0.1, 0.2]$     & Almost impossible  & It is almost impossible. \\
$(0.2, 0.3]$     & Unlikely           & It is unlikely. \\
$(0.3, 0.4]$     & Somewhat doubtful  & It is somewhat doubtful. \\
$(0.4, 0.55]$    & Possible           & It is possible. \\
$(0.55, 0.70]$   & Likely             & It is likely. \\
$(0.70, 0.80]$   & Quite likely       & It is quite likely. \\
$(0.80, 0.90]$   & Very likely        & It is very likely. \\
$(0.90, 1.00]$   & Certain            & It is certain. \\
\bottomrule
\end{tabular}
\caption{Mapping from confidence values to uncertainty hedges and postfix phrases, informed by the meta-analysis of \citet{vogel2022interpretation} on verbal–numerical correspondences. We also provide their full set of correspondences in \cref{app:vogel}.}
\label{tab:conf_to_hedger}
\end{table}

\paragraph{Hedging.}
In step (4), we map the numerical confidence value $C(a)$ of an assertion $a$ to a linguistic expression of uncertainty. For that, we draw from %
work by \citet{vogel2022interpretation}, defining a mapping that reflects common human interpretations (see \cref{tab:conf_to_hedger}). %
Depending on the strategy, this mapping yields either an uncertainty hedge or a postfix phrase. We then instantiate step (4) of faithful response generation above in one of two ways:
\begin{enumerate}[itemsep=0pt]
    \item \textbf{FUT-interweave.} Use an auxiliary language model to rewrite $a$ so that the uncertainty hedge is integrated naturally into assertion $a$. 
    \item \textbf{FUT-postfix.} Append the corresponding postfix phrase to the end of the assertion $a$.
\end{enumerate}
As for step (5), we simply concatenate hedged assertions.
Interweaving arguably produces the most natural text, but requires additional LM calls during training data generation. Postfixing, by contrast, is inexpensive as it can be applied using a template, and it preserves the original response $y$ more closely by only adding additional tokens at the end. We evaluate both strategies in \cref{sec:experiments}.

\paragraph{Belief preservation.} The faithful response generation procedure---denote it by $f(y)$---in fact identifies a distribution known in probability theory as the \emph{pushforward} of $p$ by $f$ \citep{schilling2017measures}, denoted by $p_{\#f}$. The pushforward is the distribution we get when we  `relabel' the outcomes of a random variable ($Y \sim p(\cdot|x)$ in our case) following some fixed rule or procedure ($f$ in our case). %
In effect we start with a random variable $Y \sim p(\cdot|x)$ over unhedged responses and end up with a  random variable $R\sim p_{\#f}(\cdot|x)$ over hedged responses. 
As we design $f$ to satisfy \citet{yona-etal-2024-large}'s notion of faithfulness, we obtain a distribution over faithfully hedged responses. 
Because $f$ does not modify the semantics of the assertion beyond hedging (\eg, it never changes who is blamed in the examples of \cref{fig:sketch2}), it can be shown (see \cref{app:belief-preservation}) that $p_{\#f}$ preserves the original distribution over asserted content (\eg, the probabilities of blaming  \textit{Elmo}, \textit{Grover} or \textit{Oscar} in \cref{fig:sketch2}).  %

\paragraph{Training.} The process of `relabelling'  for faithfulness (\ie, steps 1--5) can be expensive (for interactive use, for example), involving repeated sampling (for confidence estimation) and calls to various LLMs (for assertion extraction, contradiction detection and hedging). Hence, we propose to learn to reproduce its effects via maximum likelihood estimation on  data explicitly generated by $f$ (\ie, faithfully hedged and belief-preserving samples from $p_{\#f}$). This corresponds to choosing a model $q$ such that $\operatorname{KL}(p_{\#f}, q)$ is minimum. For us, $q$ is in fact a fine-tuned version of $p$.

\section{Experiments}
\label{sec:experiments}
We evaluate our method on three instruction-tuned LLMs: the 7B and 13B parameter variants of OLMo2 \citep{olmo20242} and the 8B Tülu3 Llama 3.1 model \citep{llama3,lambert2024tulu3}, starting from their instruction-tuned checkpoints. For training and evaluation we use PopQA \citep{mallen-etal-2023-trust}, a challenging open-domain QA dataset. We follow the preprocessing and subsampling as in \citet{yona-etal-2024-large} to obtain evaluation data, and use the remaining 5,654 questions after preprocessing for training data generation (see \Cref{app:datasets} for details). For each prompt we sample 10 responses and construct data under both hedging strategies, where interweaving is performed with Mistral 12B and confidence is estimated using DeBERTa Large as NLI system \citep{he2021deberta}. For efficiency's sake, we skip the assertion extraction step during synthetic data generation, as the short-form QA generations themselves predominantly concern single-assertion answers ($\geq98\%$). Data generation for both strategies took 15 hours and fine-tuning a single model took about 2.5 hours on an NVIDIA H100 GPU.%

\subsection{Faithfulness Evaluation}
\begin{table}
\footnotesize
\centering
\begin{tabular}{lccc}
\hline
\textbf{Model} & \textbf{Tülu3 8B} & \textbf{OLMo 7B} & \textbf{OLMo 13B} \\
\hline
Base (vanilla) & 0.52 & 0.53 & 0.52 \\
Base (unc.) & 0.59 & 0.55 & 0.58 \\
FUT-interweave & \textbf{0.71} & \textbf{0.79} & \textbf{0.78} \\
FUT-postfix & 0.59 & 0.71 & 0.73 \\
\hline
\end{tabular}
\caption{Faithfulness (cMFG) on PopQA for three instruction-tuned LLMs. We compare FUT (interweave, postfix) with baselines under different prompting strategies. cMFG scores of 1 indicate perfect alignment, while 0.5 indicates uncorrelated decisiveness and confidence (\cref{sec:background})}
\label{tab:popqa-faithfulness}
\end{table}

\paragraph{Faithfulness Judge.} We use GPT-5 \citep{openai2025gpt5} in our faithfulness evaluation to extract decisiveness and as NLI system for computing model confidence (\cref{sec:background}). To evaluate its reliability on these tasks, we evaluate GPT-5 on PopQA evaluation data using generations from the FUT-interweave model. We manually annotate 100 randomly sampled generations with decisiveness  labels following the verbal-numerical probability correspondences by \citet{vogel2022interpretation}.
Comparing these annotations to GPT-5's automatic decisiveness judgements yields a Spearman rank correlation of 0.92.
We further annotate 100 randomly sampled (greedy generation, sample) pairs for contradiction / no-contradiction and find GPT-5 achieves 92\% accuracy on this task.

\paragraph{Faithfulness Results.}
We evaluate faithfulness on PopQA for both the baseline models and our FUT models. 
For the baseline models, we test two prompting strategies: a rather typical brevity-inducing prompt (\texttt{vanilla}), which is also used for FUT models, and one that additionally asks for verbalised expressions of uncertainty (\texttt{uncertainty} / \texttt{unc.}); the exact prompts are provided in \cref{app:generation-prompts}.
For our method, we fine-tune both the FUT-interweave and FUT-postfix variants for each base model.
Our results are shown in \cref{tab:popqa-faithfulness}.
Under the \texttt{vanilla} prompting strategy, all base models perform close to the worst-case baseline of 0.5 faithfulness, and prompting explicitly for uncertainty hedges yields only modest improvements, consistent with the findings of \citet{yona-etal-2024-large}. 
By contrast, both FUT-interweave and FUT-postfix substantially improve upon the prompting baselines: FUT-interweave achieves the largest gains across all models, reaching up to 0.79 cMFG, while FUT-postfix also improves over the baselines but with smaller and less consistent effects. 
A possible explanation is that interwoven hedges are phrased more naturally within the flow of a response, making them easier for the language model to learn.

\begin{table}
\centering
\begin{tabular}{lccc}
\hline
\textbf{Model} & \textbf{PopQA} & \textbf{NQ} & \textbf{TriviaQA} \\
\hline
Base (vanilla) & 0.52 & 0.53 & 0.53 \\
Base (unc.) & 0.58 & 0.60 & 0.59 \\
FUT-interweave & \textbf{0.78} & \textbf{0.76} & \textbf{0.77} \\
FUT-postfix & 0.73 & 0.74 & 0.74 \\
\hline
\end{tabular}
\caption{Faithfulness (cMFG) of OLMo2 13B on PopQA, NQ, and TriviaQA, comparing base model prompting with FUT (interweave, postfix).}
\label{tab:olmo2-13b-faithfulness}
\end{table}

\paragraph{Out-of-Domain Performance.} To assess generalisation beyond PopQA, we evaluate FUT-interweave and FUT-postfix on Natural Questions \citep[NQ;][]{kwiatkowski-etal-2019-natural} and TriviaQA \citep{joshi-etal-2017-triviaqa}. For NQ, we follow the preprocessing of \citet{yona-etal-2024-large}, while for TriviaQA we subsample 932 questions from the test set to match the size of other evaluation sets (details in \cref{app:datasets}). While not strictly out-of-domain, they differ substantially in style and difficulty, making them a useful test of robustness. \cref{tab:olmo2-13b-faithfulness} shows that the gains observed by OLMo2 13B on PopQA transfer consistently to NQ and TriviaQA, with FUT-interweave again performing strongest.

\begin{table*}[t]
\centering
\footnotesize

\begin{tabularx}{\textwidth}{@{} l >{\raggedright\arraybackslash}X l >{\raggedright\arraybackslash}X @{}}
\toprule
\multicolumn{2}{l}{\textbf{(A)} \emph{Who was the director of ``Anything Else''?}} &
\multicolumn{2}{l}{\textbf{(B)} \emph{Who was the director of ``Two of Us''?}} \\
\midrule
\textbf{Greedy (vanilla)} &
The director of ``Anything Else'' was Woody Allen. &
\textbf{Greedy (vanilla)} &
Bruce Beresford directed ``Two of Us.'' \\
\rowcolor{gray!10}
\textbf{Greedy (uncertainty)} &
Woody Allen directed ``Anything Else,'' though I'm uncertain if he was the sole director or if he shared credit with others. &
\textbf{Greedy (uncertainty)} &
The director of ``Two of Us'' is John Henderson, according to IMDb, but I cannot confirm this with absolute certainty. \\
\textbf{Greedy (FUT-interweave)} &
It is very likely that Woody Allen was the director of Anything Else. &
\textbf{Greedy (FUT-interweave)} &
It is unlikely that Bryan Elsley was the director of Two of Us. \\
\midrule
\multicolumn{2}{l}{\textbf{Samples (FUT-interweave)}} &
\multicolumn{2}{l}{\textbf{Samples (FUT-interweave)}} \\
\midrule
\multicolumn{4}{@{}c@{}}{%
  \begingroup
  \setlength{\tabcolsep}{6pt}%
  \rowcolors{1}{white}{gray!10}
  \begin{tabularx}{\textwidth}{@{} >{\raggedright\arraybackslash}X >{\raggedright\arraybackslash}X @{}}
    It is certain that \textcolor{name1}{Woody Allen} was the director of Anything Else. &
    It is unlikely that \textcolor{name2}{Bryan Elsley} was the director of Two of Us. \\
    It is very likely that \textcolor{name1}{Woody Allen} was the director of Anything Else. &
    It is almost impossible that \textcolor{name3}{Roger Michell} directed Two of Us. \\
    It is almost impossible that the director of Anything Else is \textcolor{name6}{Hartman Genus}. &    
    It is unlikely that \textcolor{name2}{Bryan Elsley} was the director of Two of Us. \\
    It is very likely that \textcolor{name1}{Woody Allen} directed Anything Else. &
    It is somewhat doubtful that Two of Us was directed by \textcolor{name9}{David Burrows}. \\
    It is quite likely that \textcolor{name1}{Woody Allen} was the director of Anything Else. &
    It is unlikely that \textcolor{name10}{Penny Marshall} was the director of Two of Us. \\
  \end{tabularx}
  \endgroup
} \\
\bottomrule
\end{tabularx}

\caption{Example generations from OLMo2 13B for two questions from the PopQA evaluation set. We show greedy generations from the base model under different prompts (\texttt{vanilla} and \texttt{uncertainty}) alongside those from the FUT-interweave variant. We additionally display unbiased samples from FUT-interweave, demonstrating that hedging behaviour emerges across the distribution rather than being limited to to the greedy output. We colour samples by semantic equivalence clusters.}
\label{tab:samples}
\vspace{-0.5em}
\end{table*}

\paragraph{Example Generations.} 
In \cref{tab:samples} we show example generations from OLMo2 13B using both \texttt{vanilla} and \texttt{uncertainty} prompts, as well as its FUT-interweave variant. The \texttt{vanilla} prompt on the base model tends to produce decisive responses without hedging, whereas the \texttt{uncertainty} prompt introduces hedges that, as our cMFG evaluation shows, are not faithfully aligned with the model's uncertainty. In contrast, FUT-interweave naturally integrates uncertainty markers into its answers, with phrasing that is broadly consistent with the model’s distribution over semantically distinct responses. Importantly, hedging is not confined to the greedy output: samples from FUT-interweave also exhibit appropriate uncertainty expressions, indicating that FUT introduces faithful hedging to the full distribution rather than merely altering its top-ranked responses. We return to this point in \cref{sec:decoding-analsyis}.
\begin{table}[h!]
\centering
\begin{tabular}{lccc}
\hline
\textbf{Model} & \textbf{PopQA} & \textbf{NQ} & \textbf{TriviaQA} \\
\hline
Base (vanilla) & 23.6\% & 44.7\% & 66.1\% \\
Base (unc.) & 24.9\% & 48.0\% & 69.2\% \\
FUT-interweave & 21.9\% & 41.4\% & 65.7\% \\
FUT-postfix & 23.1\% & 41.1\% & 64.9\% \\
\hline
\end{tabular}
\caption{QA performance of OLMo2 13B: lenient exact match accuracy on PopQA, NQ, and TriviaQA for base model (\texttt{vanilla} and \texttt{unc.} prompts) vs. FUT (interweave, postfix).}
\label{tab:olmo2-13b-accuracy}
\end{table}

\subsection{Preserving Model Beliefs}
\paragraph{Impact on QA Performance.}
When fine-tuning models for improved faithfulness, our goal is to alter the model as little as possible beyond incorporating the correct uncertainty hedges. 
In \cref{tab:olmo2-13b-accuracy} we report QA performance of all variants of the OLMo2 13B model on PopQA, NQ, and TriviaQA. 
Because traditional exact match scores are difficult to compute in the presence of uncertainty hedges, we adopt a lenient variant of exact match accuracy, checking only for the presence of any reference answer in the generated string. 
We find that faithful uncertainty tuning leads to only minor reductions in QA accuracy across datasets, with performance decreasing by just 0.5–3.6 percentage points. 
This suggests that FUT successfully improves faithfulness while largely preserving the base model’s task competence.

\begin{figure}[h]
  \centering
  \includegraphics[width=0.45\textwidth]{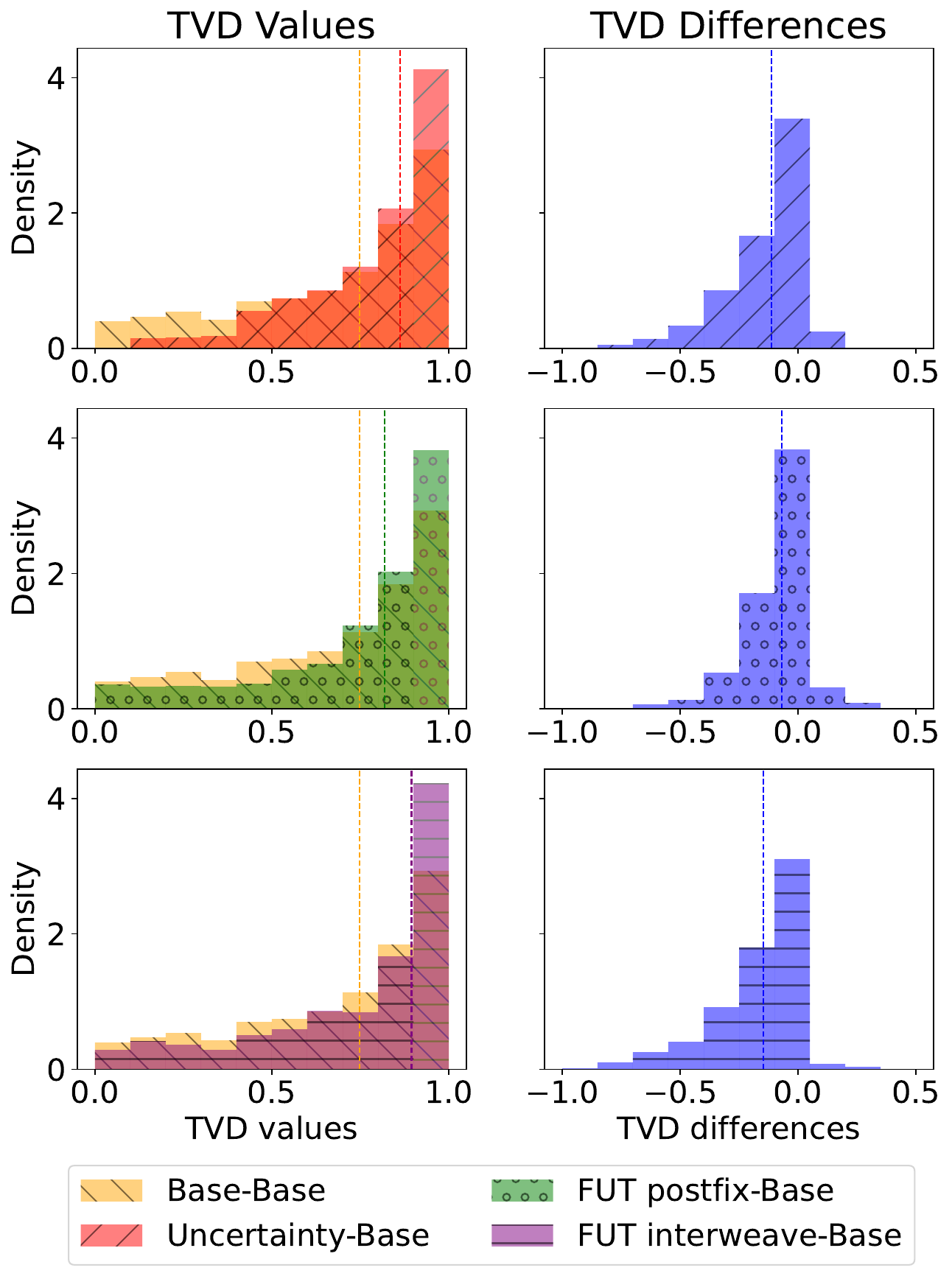}
  \hfill
    \caption{Total variation distance (TVD) between distributions over semantically distinct response clusters for OLMo2 13B. 
    Shown are TVD values for different model–prompt comparisons (left) and pairwise differences of these values (right).}
      \label{fig:tvd_main_combined}
\end{figure}

\paragraph{Semantic Distribution Shift.}
\label{sec:semantic-distribution-shift}
We next examine the \textit{semantic distribution shift} that may arise from fine-tuning: the degree to which the distribution over semantically distinct answers changes, irrespective of uncertainty hedges or surface phrasing. For each PopQA question, we sample 20 responses under different model-prompt conditions. The pooled collection of 40 responses is grouped into semantic clusters \citep{kuhn2023semantic} using LLM-based pairwise semantic equivalence judgements (details in \cref{app:semantic-distribution-shift}), where each cluster corresponds to a semantically distinct answer. This induces, for each model-prompt condition, a probability distribution over semantic clusters given by the relative frequency of responses assigned to each cluster. We then compute the total variation distance (TVD) between pairs of distributions to quantify semantic distribution shift.
We compute semantic distribution shift for OLMo2 13B under several model-prompt comparisons, shown in \cref{fig:tvd_main_combined}. As a baseline, we compare two independent sets of samples from the base model with the \texttt{vanilla} prompt (Base–Base), which provides a lower bound on the semantic distribution shift we can expect.  We also compare the base model with a \texttt{vanilla} prompt with the \texttt{uncertainty} prompt (Uncertainty-Base), as well as to FUT-interweave and FUT-postdix (FUT interweave-Base and FUT postfix-Base).  Across conditions, we observe little semantic distribution shift introduced by FUT: the distributions after fine-tuning remain very close to those of the base model. 
This is further confirmed by plotting pairwise TVD differences (right column of \Cref{fig:tvd_main_combined}), which are centered close to zero. 
Interestingly, FUT-postfix induces even less semantic shift than varying prompts in the base model.

\section{Analysis}

\subsection{Linguistic vs. Numerical Uncertainty}
\label{sec:linguistic-vs-numerical}
Our models express uncertainty in natural language, but uncertainty can also be conveyed numerically. Linguistic expressions are often more natural and intuitive for humans \citep{moxey2000communicating, liu2020intuitive, dhami2022communicating}, whereas numerical expressions are typically less ambiguous and more precise \citep{dhami2022communicating, mandel2021numerically}. In this section, we train a regression model to directly predict model confidence and compare it to our verbalized uncertainty models to assess whether the choice of communication format affects performance in producing faithful uncertainty estimates across domains.
Inspired by \citet{tangunderstanding}, that exploit LLM embeddings of strings as downstream features for
metric prediction, we employ a similar method
to predict confidence estimates.
We follow the faithful response generation procedure described in \cref{sec:faithful-uncertainty-tuning} stopping at step (3) and use estimated model confidences as regression targets. A regression head is trained on mean-pooled last-layer embeddings of question-answer pairs from OLMo2 13B and a mean-squared-error loss (architecture and hyperparameters in  \cref{app:regression}). For fair comparison with verbalised uncertainty methods, we do not freeze OLMo2's parameters. %

\begin{table}[t]
\centering
\begin{tabular}{lccc}
\hline
\textbf{Model} & \textbf{PopQA} & \textbf{NQ} & \textbf{TriviaQA} \\
\hline
FUT-interweave & 0.78 & 0.76 & 0.77 \\
FUT-postfix & 0.73 & 0.74 & 0.74 \\
FUT-numerical & 0.81 & 0.78 & 0.78 \\
\hline
\end{tabular}
\caption{Faithfulness (cMFG) of linguistic FUT models (interweave, postfix) vs. numerical (numerical) on PopQA, NQ and TriviaQA.}
\label{tab:regression}
\end{table}

Across datasets, see \cref{tab:regression}, FUT-numerical achieves slightly higher scores than  verbalised models, possibly reflecting an advantage of comparing numerical predictions directly with reference values, without  an additional decisiveness extraction step. Still, FUT-numerical performs comparably to linguistic variants, indicating that one form of uncertainty expression is not much harder to learn nor less generalisable.
The strong performance of the numerical model opens the door to uses in applications that demand more precise uncertainty estimates and personalisation of uncertainty hedges to match individual users’ verbal-numerical probability correspondences \citep{ulmer2025anthropomimetic}.  We leave the exploration of these directions to future work.

\begin{table*}[]
\centering
\scalebox{0.7}{
\begin{tabular}{l|ccc|ccc|ccc}
\hline
 & \multicolumn{3}{c|}{\textbf{FUT-Interweave (\%)}} & \multicolumn{3}{c|}{\textbf{FUT-postfix (\%)}} & \multicolumn{3}{c}{\textbf{FUT-Numerical (\%)}} \\[0.1cm]
\textbf{Uncertainty Hedge} & \multicolumn{1}{c|}{\textbf{All (932)}} & \multicolumn{1}{c|}{\textbf{Low (50)}} & \textbf{Worse (94)} & \multicolumn{1}{c|}{\textbf{All (932)}} & \multicolumn{1}{c|}{\textbf{Low (196)}} & \textbf{Worse (78)} & \multicolumn{1}{c|}{\textbf{All (932)}} & \multicolumn{1}{c|}{\textbf{Low (8)}} & \textbf{Worse (120)} \\ \hline
Impossible & \multicolumn{1}{c|}{33.90} & \multicolumn{1}{c|}{26.00} & 6.40 & \multicolumn{1}{c|}{27.47} & \multicolumn{1}{c|}{37.24} & 14.10 & \multicolumn{1}{c|}{0.97} & \multicolumn{1}{c|}{0.00} & 0.00 \\
Almost impossible & \multicolumn{1}{c|}{20.30} & \multicolumn{1}{c|}{12.00} & 8.50 & \multicolumn{1}{c|}{20.82} & \multicolumn{1}{c|}{34.69} & 12.82 & \multicolumn{1}{c|}{18.45} & \multicolumn{1}{c|}{0.00} & 0.00 \\
Unlikely & \multicolumn{1}{c|}{3.00} & \multicolumn{1}{c|}{2.00} & 2.10 & \multicolumn{1}{c|}{4.29} & \multicolumn{1}{c|}{0.00} & 2.56 & \multicolumn{1}{c|}{23.07} & \multicolumn{1}{c|}{0.00} & 0.00 \\
Somewhat doubtful & \multicolumn{1}{c|}{15.90} & \multicolumn{1}{c|}{12.00} & 15.90 & \multicolumn{1}{c|}{15.13} & \multicolumn{1}{c|}{11.73} & 5.13 & \multicolumn{1}{c|}{21.78} & \multicolumn{1}{c|}{25.00} & 6.67 \\
Possible & \multicolumn{1}{c|}{7.20} & \multicolumn{1}{c|}{8.00} & 13.80 & \multicolumn{1}{c|}{7.08} & \multicolumn{1}{c|}{2.55} & 10.26 & \multicolumn{1}{c|}{23.71} & \multicolumn{1}{c|}{75.00} & 40.00 \\
Likely & \multicolumn{1}{c|}{6.40} & \multicolumn{1}{c|}{10.00} & 12.80 & \multicolumn{1}{c|}{6.97} & \multicolumn{1}{c|}{1.53} & 15.38 & \multicolumn{1}{c|}{11.37} & \multicolumn{1}{c|}{0.00} & 48.33 \\
Quite likely & \multicolumn{1}{c|}{3.00} & \multicolumn{1}{c|}{10.00} & 5.30 & \multicolumn{1}{c|}{3.22} & \multicolumn{1}{c|}{3.57} & 3.85 & \multicolumn{1}{c|}{0.64} & \multicolumn{1}{c|}{0.00} & 5.00 \\
Very likely & \multicolumn{1}{c|}{7.50} & \multicolumn{1}{c|}{16.00} & 26.60 & \multicolumn{1}{c|}{7.08} & \multicolumn{1}{c|}{2.55} & 17.95 & \multicolumn{1}{c|}{0.00} & \multicolumn{1}{c|}{0.00} & 0.00 \\
Certain & \multicolumn{1}{c|}{1.80} & \multicolumn{1}{c|}{4.00} & 7.40 & \multicolumn{1}{c|}{7.94} & \multicolumn{1}{c|}{6.12} & 17.95 & \multicolumn{1}{c|}{0.00} & \multicolumn{1}{c|}{0.00} & 0.00 \\ \hline
\end{tabular}
}
\caption{Distribution of hedges in low faithful generation instances (Low) or those with lower values than the base model (Worse). We compare proportions with those in the full PopQA evaluation set (All).}
\label{tab:problematic_instances_hedgers}
\end{table*}

\begin{table*}[]
\centering
\scalebox{0.75}{
\begin{tabular}{l|ccc|ccc|ccc}
\hline
 & \multicolumn{3}{c|}{\textbf{FUT-interweave (\%)}} & \multicolumn{3}{c|}{\textbf{FUT-postfix (\%)}} & \multicolumn{3}{c}{\textbf{FUT-numerical (\%)}} \\[0.1cm]
{\color[HTML]{1F1F1F} \textbf{Question}} & \multicolumn{1}{l|}{\textbf{All (932)}} & \multicolumn{1}{l|}{\textbf{Low (50)}} & \multicolumn{1}{l|}{\textbf{Worse (94)}} & \multicolumn{1}{c|}{\textbf{All (932)}} & \multicolumn{1}{c|}{\textbf{Low (196)}} & \textbf{Worse (78)} & \multicolumn{1}{l|}{\textbf{All (932)}} & \multicolumn{1}{l|}{\textbf{Low (8)}} & \multicolumn{1}{l}{\textbf{Worse (120)}} \\ \hline
Person & \multicolumn{1}{c|}{83.70} & \multicolumn{1}{c|}{\cellcolor[HTML]{FFFFFF}94.00} & \cellcolor[HTML]{FFFFFF}88.30 & \multicolumn{1}{c|}{83.70} & \multicolumn{1}{c|}{83.67} & 84.62 & \multicolumn{1}{c|}{83.70} & \multicolumn{1}{c|}{87.50} & \cellcolor[HTML]{FFFFFF}81.70 \\
{\color[HTML]{1F1F1F} Birth city} & \multicolumn{1}{c|}{8.00} & \multicolumn{1}{c|}{\cellcolor[HTML]{FFFFFF}4.00} & \cellcolor[HTML]{FFFFFF}8.50 & \multicolumn{1}{c|}{8.00} & \multicolumn{1}{c|}{9.69} & 10.26 & \multicolumn{1}{c|}{8.00} & \multicolumn{1}{c|}{0.00} & \cellcolor[HTML]{FFFFFF}3.30 \\
Occupation & \multicolumn{1}{c|}{8.30} & \multicolumn{1}{c|}{\cellcolor[HTML]{FFFFFF}2.00} & \cellcolor[HTML]{FFFFFF}3.20 & \multicolumn{1}{c|}{8.30} & \multicolumn{1}{c|}{6.63} & 5.13 & \multicolumn{1}{c|}{8.30} & \multicolumn{1}{c|}{12.50} & \cellcolor[HTML]{FFFFFF}15.00 \\ \hline
\end{tabular}
}

\caption{Distribution of question types in low faithfulness instances (Low) or those with lower values than the base model (Worse). We compare proportions with those in the full PopQA evaluation set (All).}

\label{tab:problematic_instances_question_types}
\end{table*}

\subsection{Error Analysis}

\begin{figure}
  \centering
  \includegraphics[width=0.45\textwidth]{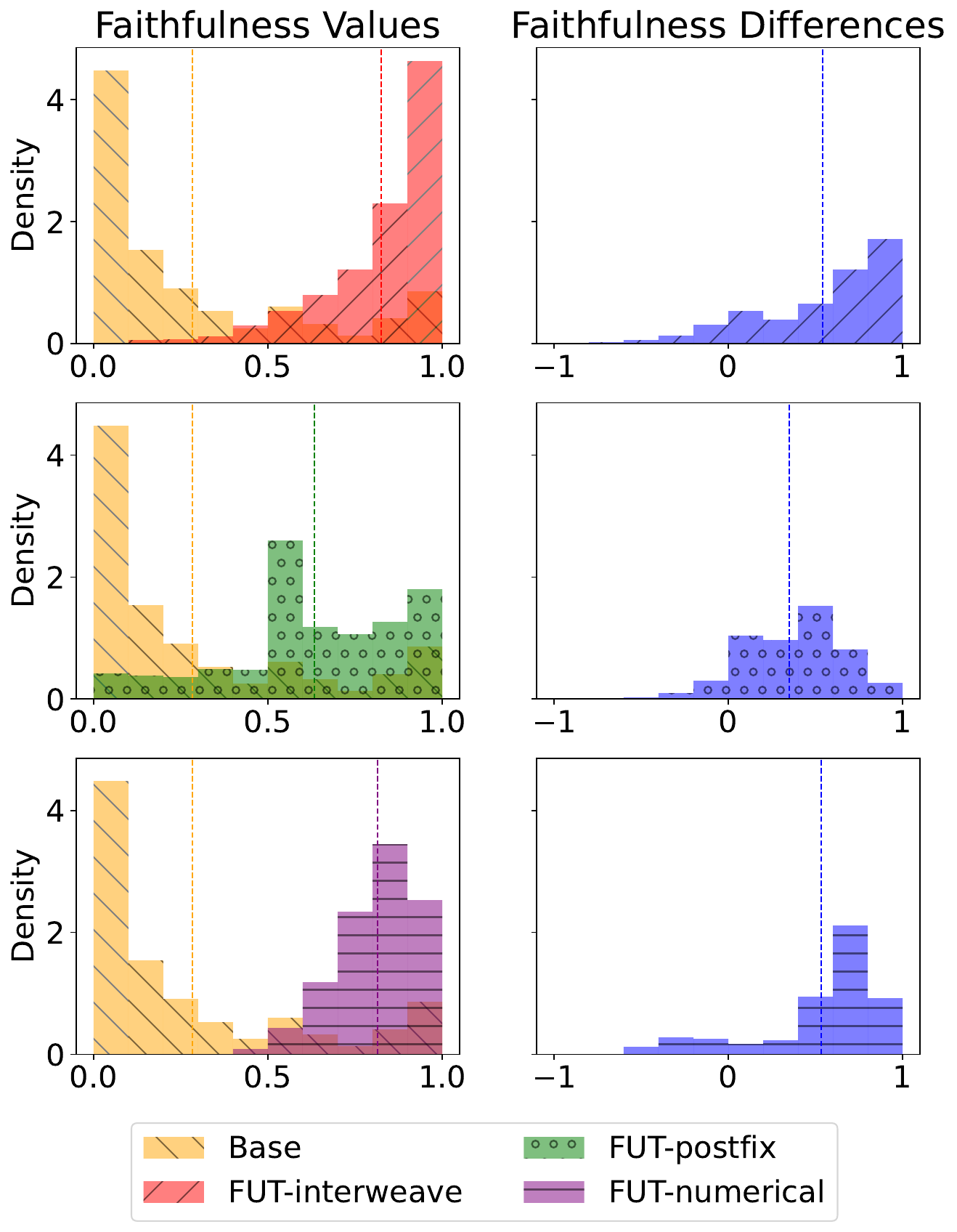}
  \hfill
  \caption{Distribution of faithful generation scores (left) for OLMo2 13B variants: the base model (\texttt{vanilla} prompt, Base), and FUT (interweave, postfix, and numerical). %
  We also plot the instance-level differences from the Base model (right).
  }
  \label{fig:faith_dist}
\end{figure}

FUT models outperform their non fine-tuned counterparts in faithful hedging. We plot instance-level faithfulness on PopQA for OLMo2 13B using the \texttt{vanilla} prompt (Base) and FUT (interweave, postfix, and numerical), see \Cref{fig:faith_dist}, and observe a shift of FUT models towards higher faithfulness values, with FUT-postfix showing less substantial gains than the other methods. %

\begin{figure*}[h]
    \begin{subfigure}{0.33\textwidth}
        \centering
        \includegraphics[width=\linewidth]{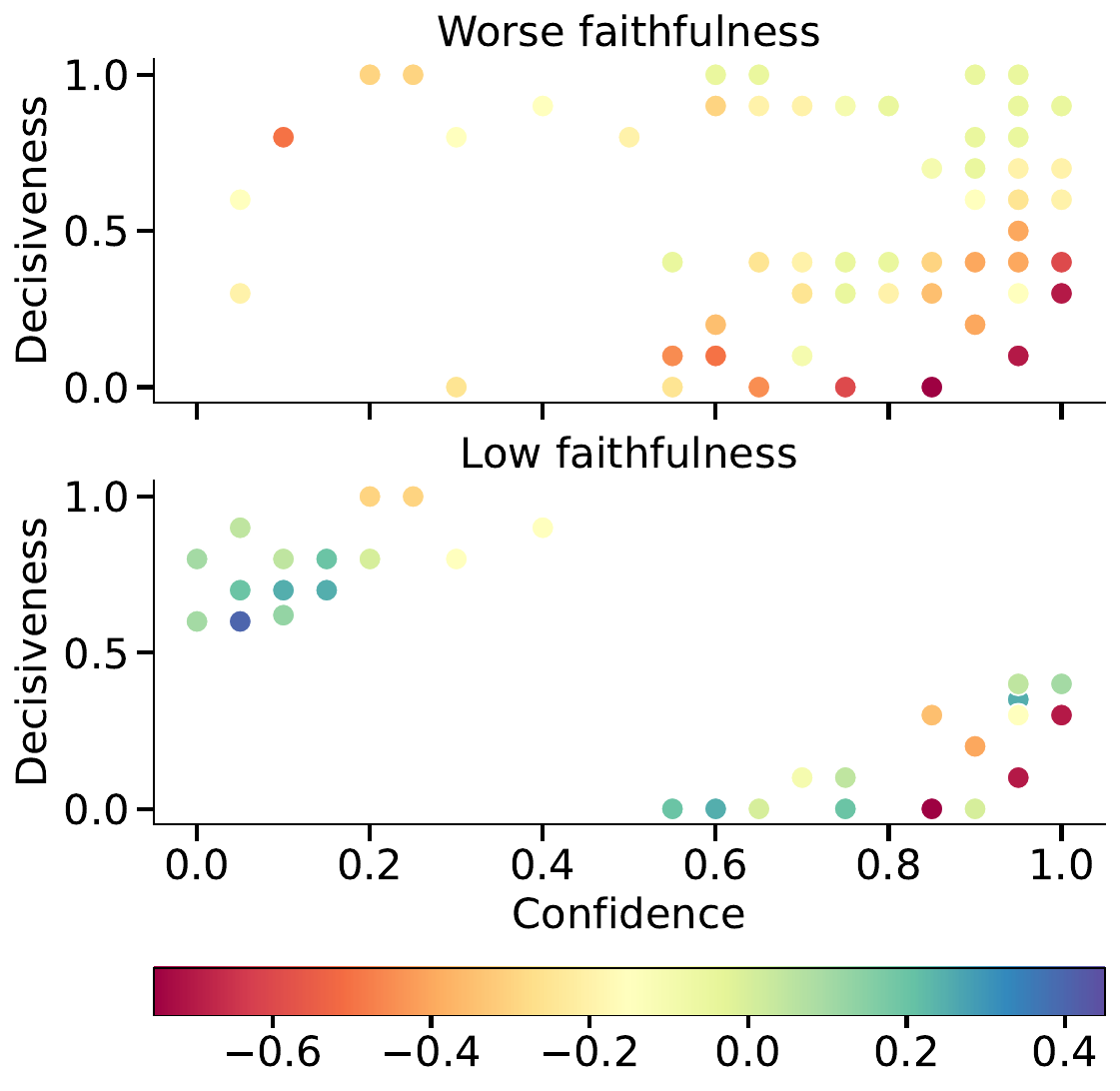}
        \caption{FUT-Interweave}
    \end{subfigure}%
    \begin{subfigure}{0.33\textwidth}
        \centering
        \includegraphics[width=\linewidth]{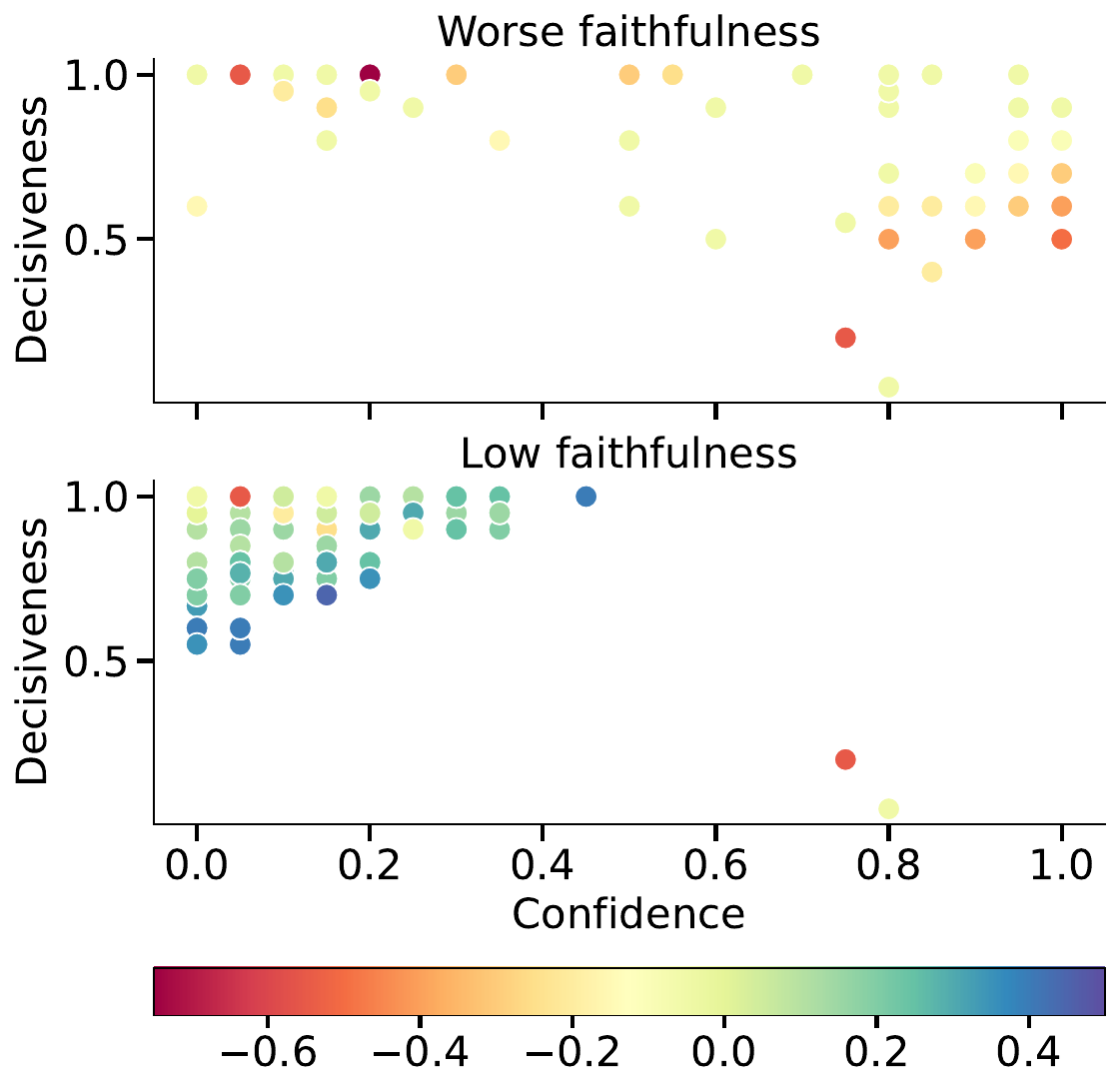}
        \caption{FUT-Postfix}
    \end{subfigure}%
    \begin{subfigure}{0.33\textwidth}
        \centering
        \includegraphics[width=\linewidth]{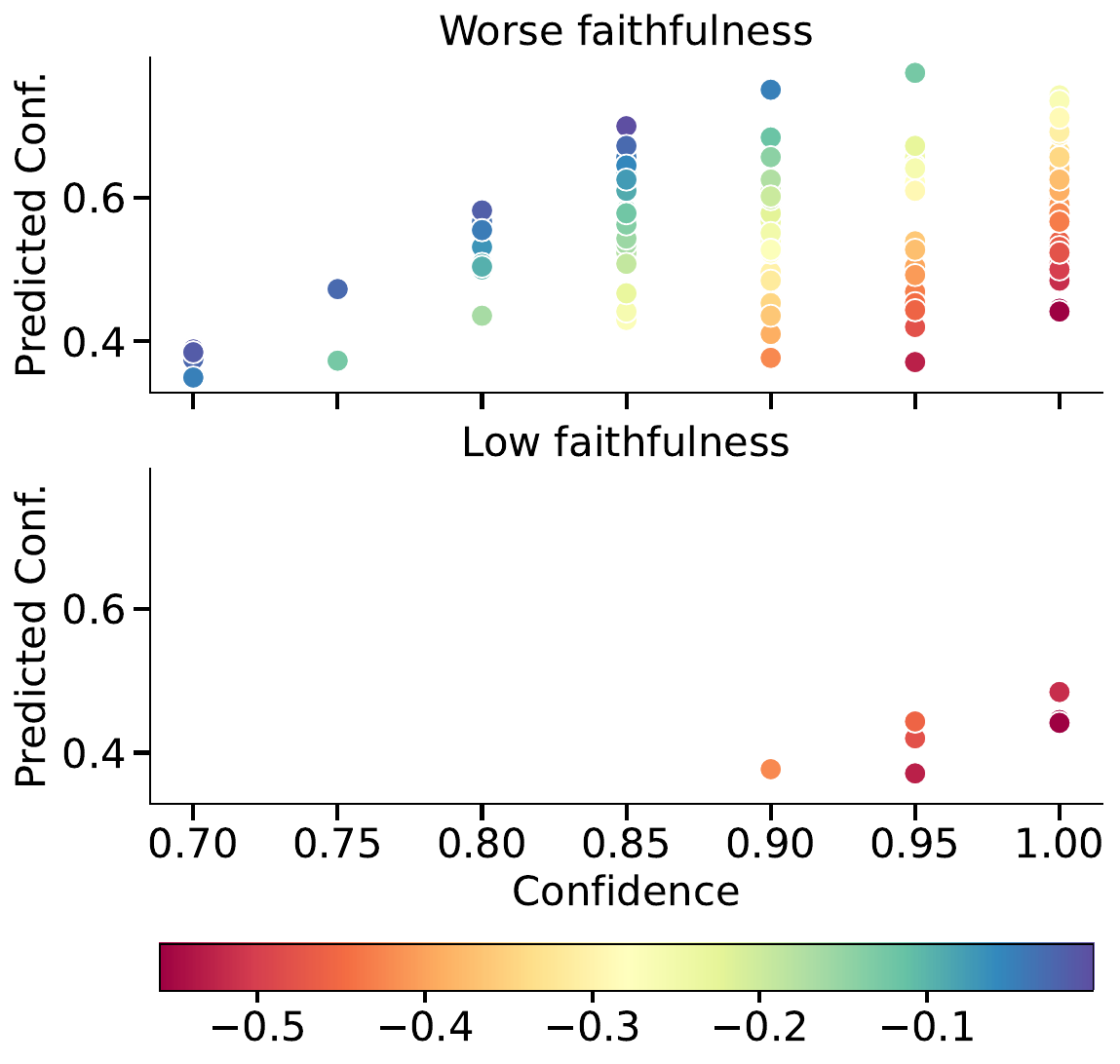}
        \caption{FUT-Numerical}
    \end{subfigure}%
    \caption{Scatter plots of confidence vs. decisiveness for failure cases: instances with reduced faithful generation values after fine-tuning (top) and those below 0.5 (bottom). Colours show changes relative to the base OLMo2 13B (\texttt{vanilla} prompt), with negative values indicating drops and vice versa.}
        
  \label{fig:faith_scatter}
\end{figure*}

To better understand shortcomings, we analyse instances of question-answer pairs that were deemed as `failure' cases under two definitions: \textit{i)} instances for which FUT models' faithfulness is lower than the base model (Worse set), and  \textit{ii)} instances for which FUT models' faithfulness is low, thresholded at values less than 0.5 (Low Set). In \cref{fig:faith_scatter} we present scatter plots of `failure' cases for FUT models. We observe that for FUT-interweave and FUT-numerical the largest performance drops occur for questions where the model has high confidence, pointing towards some issues of `under-hedging' (\eg, FUT-numerical predicts lower confidences, and FUT-interweave generates answers that are less decisive than they should).
FUT-postfix's performance drops, although smaller, seem to occur across varying confidence levels. For the Low set of instances of FUT-interweave and FUT-postfix, despite their low faithfulness, a large fraction has improved compared to the base model (see positive changes in faithfulness values,  represented by the  colour of the markers in \Cref{fig:faith_scatter}, bottom plot). For FUT-numerical, although no instances with improvements were observed, the Low set contains only a handful of instances.

We turn to analyse features of such `failure' cases and study how often these occur in these problematic sets (Low, Worse) compared to the entire evaluation set (\cref{tab:problematic_instances_hedgers}). 
For FUT-interweave, some middle and high decisiveness hedges appear considerably more often in the Worse set than in the entire evaluation set.
A similar phenomenon is observed in the Low set, mostly for high decisiveness hedges, with phrases like ``It is very likely'' or even ``It is certain'' becoming over-represented. 
Such phrasing overstates the model's certainty, leading to a mismatch between decisiveness and probabilistic belief.
By contrast, very low decisiveness hedges (``Impossible'', ``Almost impossible'') are much less frequent in the problematic sets, suggesting that these are applied more appropriately. %
For FUT-postfix, the majority of low faithfulness instances appear in the low decisiveness hedges, while decreases in faithfulness after fine-tuning tend to occur for high decisiveness hedges. 
Finally, FUT-numerical does not produce problematic cases in the extreme bins, but frequently misplaces predictions in the middle ranges, %
indicating difficulty with faithfully handling mid-confidence cases. %

Similarly, we examine the distribution of question types in the problematic sets (\cref{tab:problematic_instances_question_types}). Although differences are modest, we observe person-identification questions occurring slightly more often in problematic sets for FUT-interweave. By contrast, FUT-numerical shows the opposite trend, with failures more concentrated on birth city and occupation questions, suggesting this model may find such categories harder to handle faithfully. For FUT-postfix, the distribution of problematic instances closely mirrors that of the entire set, indicating no error association with question types.

\subsection{The Choice of Uncertainty Hedges}
In our main experiments, we fine-tuned models using the mapping of confidence values to uncertainty hedges in \cref{tab:conf_to_hedger}.
However, this mapping is not unique, and alternative sets of hedges could plausibly influence model behaviour.
To examine the sensitivity of FUT to this design choice, we test an alternative mapping of confidence to hedges. Specifically, we replace the uncertainty hedge in each bin in \cref{tab:conf_to_hedger} from low to high confidence with: highly improbable, improbable, doubtful, uncertain, maybe, reasonably possible, good chance, highly probable, and definite. 
We also experiment with a stochastic variant that randomly alternates between both mappings during data generation. To ensure consistency in faithfulness evaluation, we adapt the decisiveness extraction prompt in \cref{tab:judge-prompts} by replacing each original hedge example with its corresponding hedge from the alternative mapping.
We train OLMo2 13B using FUT-interweave on PopQA and find that both the alternative and stochastic variants achieve a cMFG of 0.80, slightly higher than the original mapping (0.78), indicating that FUT is robust to the reasonable choices of uncertainty hedges.

\subsection{Choice of Decoding Strategy}
\label{sec:decoding-analsyis}
Our faithfulness evaluation has focused on greedy decoding, following \citet{yona-etal-2024-large}, which is a standard choice, but ultimately arbitrary. Our fine-tuning procedure hedges the model’s full distribution by training on its own unbiased samples which were themselves augmented with uncertainty hedges. To assess robustness to this choice, we evaluate the faithfulness of unbiased samples and nucleus samples \citep{Holtzman2020The}. We use the FUT-interweave variant of OLMo2 13B and evaluate on PopQA. Unbiased sampling achieves an impressive 0.76 cMFG score, only slightly lower than greedy decoding (0.78), while nucleus sampling ($p=0.7$) achieves a cMFG of 0.80. This confirms that FUT improves faithfulness not only for the top-ranked outputs, but also across the wider distribution of model generations.%

\section{Related Work}

\paragraph{Numerical Confidence.} Prior work on uncertainty quantification in language models has commonly focused on producing numerical confidence estimates. Approaches include prompting models to self-report a numeric confidence score through next-token prediction alongside their answer \citep{kadavath2022language, yangverbalized, jang2025self}, training post-hoc probes on model representations that regress or classify correctness  \citep{mahaut2024factual, beigi2024internalinspector, cencerrado2025no}, and training or fine-tuning language models or auxiliary predictors to generate calibrated confidence scores through next-token prediction  \citep{ulmer-etal-2024-calibrating, han2024enhancing, krishnan2024enhancing,han2025mind,li2025conftuner,li2025graphbased}.

\paragraph{Linguistic Confidence.}
A growing body of work explores how language models can communicate their confidence directly in natural language rather than numerically. \citet{linteaching} first show that GPT-3 can be trained to produce additional tokens indicative of empirical correctness of the response (\eg ``C: high confidence''). \citet{mielke2022reducing} instead train a correctness predictor (``calibrator'') and use controllable generation to include an appropriate uncertainty hedge into the response. \citet{xu2024sayself} fine-tune language models on LLM-generated summaries of disagreements across reasoning chains, so that the resulting models produce calibrated uncertainty expressions directly in natural language. 
\citet{band2024linguistic} also leverage sample consistency with LLM-generated summarisations, but extend this approach to long-form generations containing multiple factual statements, combining supervised fine-tuning with reinforcement learning to ensure that interwoven hedges reflect empirical correctness. Most similar to our work, \citet{chaudhry2024finetuning} map model-estimated probabilities of correctness into human-interpretable hedges and fine-tune models on synthetic data where these hedges are interwoven naturally into the responses.

\paragraph{Faithfulness.}
Crucially, all the methods mentioned above are concerned with \emph{calibration}, \ie ensuring numerical or linguistic expressions of confidence align with the empirical probability of correctness. This perspective ties predictions to dataset-level error rates rather than to the model’s own uncertainty on a given answer, so models may appear calibrated while still being misaligned with their actual beliefs. 
In recent years, several metrics have been proposed to capture this \emph{faithfulness gap} between what the models expresses and the uncertainty reflected in its own distribution. For example, comparing token probabilities with the decisiveness of model assertions \citep{kumar2024confidence}, or assessing whether string summaries of the internal answer distribution faithfully reflect that distribution 
\citep{kirchhof2025self}. \citet{yona-etal-2024-large} introduce the conditional mean faithful generation metric (cMFG; \cref{sec:background}), which measures how well a model-based notion of confidence based on sample consistency aligns with the decisiveness of its verbalised assertions, showing that even strong commercial LLMs exhibit low faithfulness. In the same spirit, \citet{ji2025calibrating} study misalignment between internal model uncertainty and expressed decisiveness to detect hallucinations. 
As our adopted uncertainty heuristic is most consistent with the faithfulness framework proposed in \citet{yona-etal-2024-large}, we choose to adopt cMFG as our evaluation framework of choice. Also adopting this notion of faithfulness, \citet{liu2025metafaith} propose \textsc{MetaFaith}, a black-box prompting method that instructs models to explicitly reflect on their own confidence when generating answers. While this yields slight improvements in cMFG over standard prompting, typical values remain modest; by contrast, our fine-tuning approach points to larger gains.

\section{Discussion}
Our results show that Faithful Uncertainty Tuning substantially reduces the gap between models’ internal probabilistic uncertainty and the decisiveness of their verbal output. Across multiple models and QA datasets, FUT achieves large gains in faithfulness with only minimal semantic distribution shift and without compromising task accuracy.  This aligns with our design principle of separating concerns: initial model training is responsible for developing accurate beliefs about the world, while our lightweight fine-tuning method ensures that these beliefs are communicated faithfully. %
We find our method to be robust to the choice of hedges, decoding strategy, and even numerical expressions of uncertainty.

In our framing, correctness is the responsibility of the base instruction-tuned model (or upstream improvements such as continued pretraining, retrieval-augmented generation, or alignment), and FUT is a lightweight post-hoc adaptation that makes the model’s beliefs explicit. This contrasts with calibration-oriented fine-tuning, which requires supervision about correctness and adapts the model’s distribution toward empirical accuracy. While valuable in applications that demand actionable risk estimates, it comes at the cost of injecting new task knowledge and potentially discarding the model’s original beliefs. FUT does not attempt to relearn correctness, but instead surfaces uncertainty already implicit in the model. If such knowledge is present, it should be communicated faithfully rather than obscured, and our results show that FUT successfully achieves this.%

Overall, we show that FUT surfaces model uncertainty without altering underlying beliefs. Applying it beyond QA, \textit{e.g.} to long-form generation, will require careful tuning of claim segmentation and hedge strategies, but our results establish a clear foundation for such future developments.

\section*{Limitations}
While our approach enables models to verbalise their uncertainty in a way that is faithful to their own beliefs, it does not prevent them from being confidently wrong. FUT surfaces uncertainty already present in the base model, but if the underlying distribution is misinformed or overconfident, then faithfully verbalised responses may still mislead users.

Our evaluation of faithfulness relies on the use of the use of LLMs for both extracting decisiveness and performing natural language inference to compute confidence scores. Although we validated their reliability through manual checks, these judgements may introduce biases or errors that affect reported scores.
Moreover, we use OpenAI's GPT-5 model for these tasks, thus making the evaluation setup costly: evaluating a single model-dataset combination (of about 1,000 questions) using 20 samples for computing confidence scores (\textit{i.e.} doing 20 NLI inferences) plus decisiveness extraction of the greedy response costs about \$25 in OpenAI API credits.
While this is somewhat manageable for research-scale experiments, it may hinder large-scale reproducibility. 
In principle, cheaper locally run models could replace GPT-5 in our setup, but for comparability with prior work \citep{yona-etal-2024-large,yona2024can} we opted to follow a similar evaluation protocol.

Finally, our experiments are limited to open-domain QA. It remains to be seen how FUT generalises to other tasks such as long-form generation, dialogue, or multi-claim summarisation. %

\section*{Acknowledgements}
\begin{wrapfigure}[2]{l}{0.12\linewidth}
\vspace{-15pt}
\includegraphics[width=0.08\textwidth]{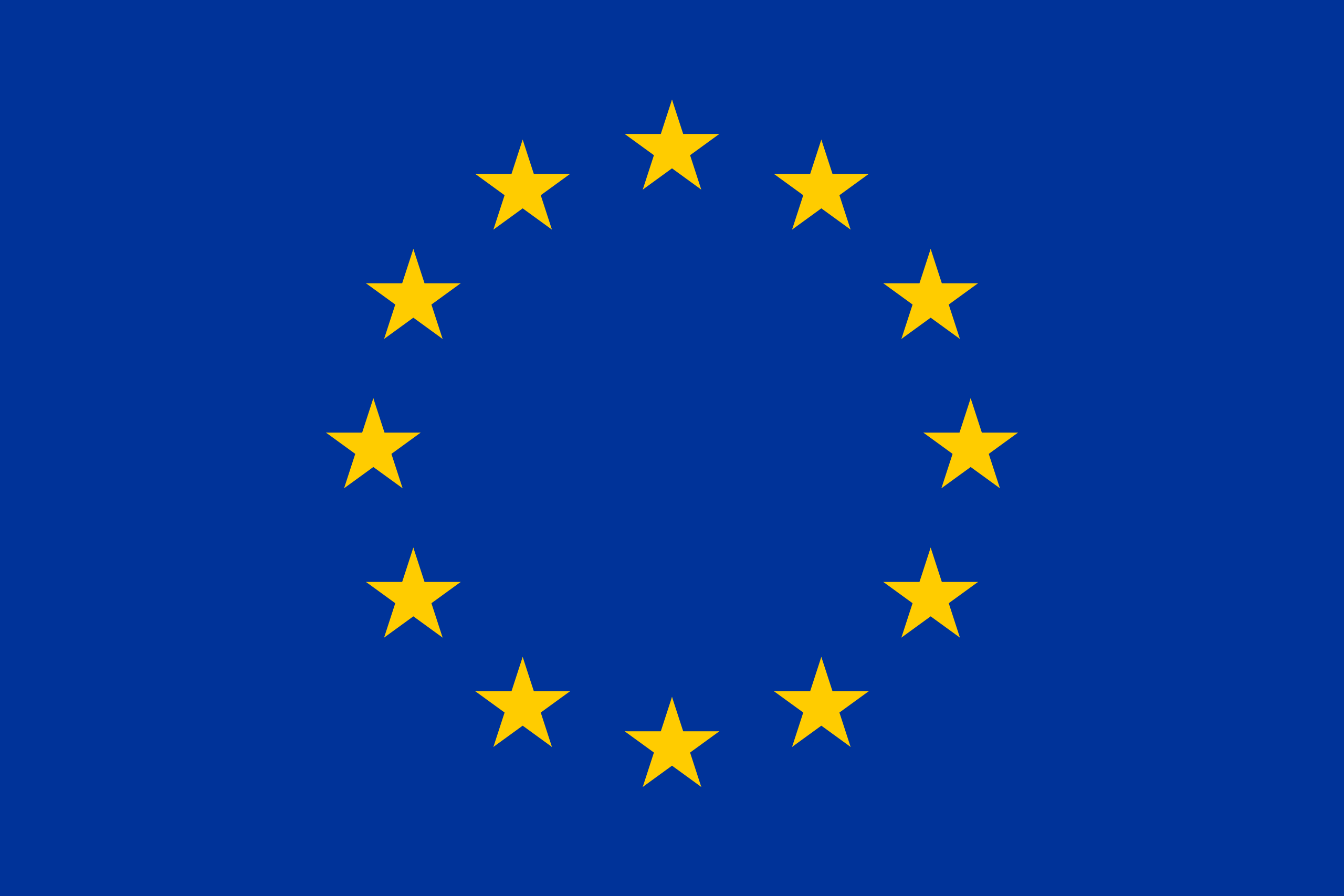}
\end{wrapfigure} 
This project received funding from the European Union’s Horizon Europe Research and Innovation programme under Grant Agreement No 101070631 (UTTER). We thank Joris Baan and Dennis Ulmer for their valuable suggestions and insightful discussions on this project.

\bibliography{custom,anthology}

\begin{thebibliography}{52}
\expandafter\ifx\csname natexlab\endcsname\relax\def\natexlab#1{#1}\fi

\bibitem[{Band et~al.(2024)Band, Li, Ma, and Hashimoto}]{band2024linguistic}
Neil Band, Xuechen Li, Tengyu Ma, and Tatsunori Hashimoto. 2024.
\newblock Linguistic calibration of long-form generations.
\newblock In \emph{International Conference on Machine Learning}, pages 2732--2778. PMLR.

\bibitem[{Beigi et~al.(2024)Beigi, Shen, Yang, Lin, Wang, Mohan, He, Jin, Lu, and Huang}]{beigi2024internalinspector}
Mohammad Beigi, Ying Shen, Runing Yang, Zihao Lin, Qifan Wang, Ankith Mohan, Jianfeng He, Ming Jin, Chang-Tien Lu, and Lifu Huang. 2024.
\newblock Internalinspector i: Robust confidence estimation in llms through internal states.
\newblock \emph{CoRR}.

\bibitem[{Cencerrado et~al.(2025)Cencerrado, Masdemont, Hawthorne, Africa, and Pacchiardi}]{cencerrado2025no}
Iv{\'a}n Vicente~Moreno Cencerrado, Arnau~Padr{\'e}s Masdemont, Anton~Gonzalvez Hawthorne, David~Demitri Africa, and Lorenzo Pacchiardi. 2025.
\newblock No answer needed: Predicting llm answer accuracy from question-only linear probes.
\newblock \emph{arXiv preprint arXiv:2509.10625}.

\bibitem[{Chaudhry et~al.(2024)Chaudhry, Thiagarajan, and Gorur}]{chaudhry2024finetuning}
Arslan Chaudhry, Sridhar Thiagarajan, and Dilan Gorur. 2024.
\newblock Finetuning language models to emit linguistic expressions of uncertainty.
\newblock \emph{arXiv preprint arXiv:2409.12180}.

\bibitem[{Dettmers et~al.(2022)Dettmers, Lewis, Shleifer, and Zettlemoyer}]{dettmers2022optimizers}
Tim Dettmers, Mike Lewis, Sam Shleifer, and Luke Zettlemoyer. 2022.
\newblock 8-bit optimizers via block-wise quantization.
\newblock \emph{9th International Conference on Learning Representations, ICLR}.

\bibitem[{Dhami and Mandel(2022)}]{dhami2022communicating}
Mandeep~K Dhami and David~R Mandel. 2022.
\newblock Communicating uncertainty using words and numbers.
\newblock \emph{Trends in Cognitive Sciences}, 26(6):514--526.

\bibitem[{Dubey et~al.(2024)Dubey, Jauhri, Pandey, Kadian, Al-Dahle, Letman, Mathur, Schelten, Yang, Fan, Goyal, Hartshorn, Yang, Mitra, Sravankumar, Korenev, Hinsvark, Rao, Zhang, Rodriguez, Gregerson, Spataru, Rozière, Biron, Tang, Chern, Caucheteux, Nayak, Bi, Marra, McConnell, Keller, Touret, Wu, Wong, Ferrer, Nikolaidis, Allonsius, Song, Pintz, Livshits, Esiobu, Choudhary, Mahajan, Garcia-Olano, Perino, Hupkes, Lakomkin, AlBadawy, Lobanova, Dinan, Smith, Radenovic, Zhang, Synnaeve, Lee, Anderson, Nail, Mialon, Pang, Cucurell, Nguyen, Korevaar, Xu, Touvron, Zarov, Ibarra, Kloumann, Misra, Evtimov, Copet, Lee, Geffert, Vranes, Park, Mahadeokar, Shah, van~der Linde, Billock, Hong, Lee, Fu, Chi, Huang, Liu, Wang, Yu, Bitton, Spisak, Park, Rocca, Johnstun, Saxe, Jia, Alwala, Upasani, Plawiak, Li, Heafield, Stone, and et~al.}]{llama3}
Abhimanyu Dubey, Abhinav Jauhri, Abhinav Pandey, Abhishek Kadian, Ahmad Al-Dahle, Aiesha Letman, Akhil Mathur, Alan Schelten, Amy Yang, Angela Fan, Anirudh Goyal, Anthony Hartshorn, Aobo Yang, Archi Mitra, Archie Sravankumar, Artem Korenev, Arthur Hinsvark, Arun Rao, Aston Zhang, Aurélien Rodriguez, Austen Gregerson, Ava Spataru, Baptiste Rozière, Bethany Biron, Binh Tang, Bobbie Chern, Charlotte Caucheteux, Chaya Nayak, Chloe Bi, Chris Marra, Chris McConnell, Christian Keller, Christophe Touret, Chunyang Wu, Corinne Wong, Cristian~Canton Ferrer, Cyrus Nikolaidis, Damien Allonsius, Daniel Song, Danielle Pintz, Danny Livshits, David Esiobu, Dhruv Choudhary, Dhruv Mahajan, Diego Garcia-Olano, Diego Perino, Dieuwke Hupkes, Egor Lakomkin, Ehab AlBadawy, Elina Lobanova, Emily Dinan, Eric~Michael Smith, Filip Radenovic, Frank Zhang, Gabriel Synnaeve, Gabrielle Lee, Georgia~Lewis Anderson, Graeme Nail, Grégoire Mialon, Guan Pang, Guillem Cucurell, Hailey Nguyen, Hannah Korevaar, Hu~Xu, Hugo Touvron, Iliyan Zarov,
  Imanol~Arrieta Ibarra, Isabel~M. Kloumann, Ishan Misra, Ivan Evtimov, Jade Copet, Jaewon Lee, Jan Geffert, Jana Vranes, Jason Park, Jay Mahadeokar, Jeet Shah, Jelmer van~der Linde, Jennifer Billock, Jenny Hong, Jenya Lee, Jeremy Fu, Jianfeng Chi, Jianyu Huang, Jiawen Liu, Jie Wang, Jiecao Yu, Joanna Bitton, Joe Spisak, Jongsoo Park, Joseph Rocca, Joshua Johnstun, Joshua Saxe, Junteng Jia, Kalyan~Vasuden Alwala, Kartikeya Upasani, Kate Plawiak, Ke~Li, Kenneth Heafield, Kevin Stone, and et~al. 2024.
\newblock \href {https://doi.org/10.48550/arXiv.2407.21783} {The llama 3 herd of models}.
\newblock \emph{CoRR}, abs/2407.21783.

\bibitem[{Farquhar et~al.(2024)Farquhar, Kossen, Kuhn, and Gal}]{farquhar2024detecting}
Sebastian Farquhar, Jannik Kossen, Lorenz Kuhn, and Yarin Gal. 2024.
\newblock Detecting hallucinations in large language models using semantic entropy.
\newblock \emph{Nature}, 630(8017):625--630.

\bibitem[{Han et~al.(2024)Han, Li, Chen, Shi, Du, Xiao, Liang, and Lin}]{han2024enhancing}
Haixia Han, Tingyun Li, Shisong Chen, Jie Shi, Chengyu Du, Yanghua Xiao, Jiaqing Liang, and Xin Lin. 2024.
\newblock Enhancing confidence expression in large language models through learning from past experience.
\newblock \emph{CoRR}.

\bibitem[{Han et~al.(2025)Han, Li, Chen, Shi, Wang, Yue, Liang, Lin, Wen, Chen et~al.}]{han2025mind}
Jinyi Han, Tingyun Li, Shisong Chen, Jie Shi, Xinyi Wang, Guanglei Yue, Jiaqing Liang, Xin Lin, Liqian Wen, Zulong Chen, et~al. 2025.
\newblock Mind the generation process: Fine-grained confidence estimation during llm generation.
\newblock \emph{arXiv preprint arXiv:2508.12040}.

\bibitem[{He et~al.(2021)He, Liu, Gao, and Chen}]{he2021deberta}
Pengcheng He, Xiaodong Liu, Jianfeng Gao, and Weizhu Chen. 2021.
\newblock \href {https://openreview.net/forum?id=XPZIaotutsD} {Deberta: Decoding-enhanced bert with disentangled attention}.
\newblock In \emph{International Conference on Learning Representations}.

\bibitem[{Holtzman et~al.(2020)Holtzman, Buys, Du, Forbes, and Choi}]{Holtzman2020The}
Ari Holtzman, Jan Buys, Li~Du, Maxwell Forbes, and Yejin Choi. 2020.
\newblock \href {https://openreview.net/forum?id=rygGQyrFvH} {The curious case of neural text degeneration}.
\newblock In \emph{International Conference on Learning Representations}.

\bibitem[{Jang et~al.(2025)Jang, Jang, Lee, Ok, and Ahn}]{jang2025self}
Hyosoon Jang, Yunhui Jang, Sungjae Lee, Jungseul Ok, and Sungsoo Ahn. 2025.
\newblock Self-training large language models with confident reasoning.
\newblock \emph{arXiv preprint arXiv:2505.17454}.

\bibitem[{Ji et~al.(2025)Ji, Yu, Koishekenov, Bang, Hartshorn, Schelten, Zhang, Fung, and Cancedda}]{ji2025calibrating}
Ziwei Ji, Lei Yu, Yeskendir Koishekenov, Yejin Bang, Anthony Hartshorn, Alan Schelten, Cheng Zhang, Pascale Fung, and Nicola Cancedda. 2025.
\newblock Calibrating verbal uncertainty as a linear feature to reduce hallucinations.
\newblock \emph{arXiv preprint arXiv:2503.14477}.

\bibitem[{Joo et~al.(2025)Joo, Min, Koo, and Jung}]{Joo2025BlackBoxHD}
Seongho Joo, Kyungmin Min, Jahyun Koo, and Kyomin Jung. 2025.
\newblock \href {https://www.arxiv.org/pdf/2509.21999} {Black-box hallucination detection via consistency under the uncertain expression}.

\bibitem[{Joshi et~al.(2017)Joshi, Choi, Weld, and Zettlemoyer}]{joshi-etal-2017-triviaqa}
Mandar Joshi, Eunsol Choi, Daniel Weld, and Luke Zettlemoyer. 2017.
\newblock \href {https://doi.org/10.18653/v1/P17-1147} {{T}rivia{QA}: A large scale distantly supervised challenge dataset for reading comprehension}.
\newblock In \emph{Proceedings of the 55th Annual Meeting of the Association for Computational Linguistics (Volume 1: Long Papers)}, pages 1601--1611, Vancouver, Canada. Association for Computational Linguistics.

\bibitem[{Kadavath et~al.(2022)Kadavath, Conerly, Askell, Henighan, Drain, Perez, Schiefer, Hatfield-Dodds, DasSarma, Tran-Johnson et~al.}]{kadavath2022language}
Saurav Kadavath, Tom Conerly, Amanda Askell, Tom Henighan, Dawn Drain, Ethan Perez, Nicholas Schiefer, Zac Hatfield-Dodds, Nova DasSarma, Eli Tran-Johnson, et~al. 2022.
\newblock Language models (mostly) know what they know.
\newblock \emph{CoRR}.

\bibitem[{Kim et~al.(2024)Kim, Liao, Vorvoreanu, Ballard, and Vaughan}]{kim2024m}
Sunnie~SY Kim, Q~Vera Liao, Mihaela Vorvoreanu, Stephanie Ballard, and Jennifer~Wortman Vaughan. 2024.
\newblock " i'm not sure, but...": Examining the impact of large language models' uncertainty expression on user reliance and trust.
\newblock In \emph{Proceedings of the 2024 ACM Conference on Fairness, Accountability, and Transparency}, pages 822--835.

\bibitem[{Kirchhof et~al.(2025)Kirchhof, F{\"u}ger, Goli{\'n}ski, Dhekane, Blaas, and Williamson}]{kirchhof2025self}
Michael Kirchhof, Luca F{\"u}ger, Adam Goli{\'n}ski, Eeshan~Gunesh Dhekane, Arno Blaas, and Sinead Williamson. 2025.
\newblock Self-reflective uncertainties: Do llms know their internal answer distribution?
\newblock \emph{arXiv preprint arXiv:2505.20295}.

\bibitem[{Krishnan et~al.(2024)Krishnan, Khanna, and Tickoo}]{krishnan2024enhancing}
Ranganath Krishnan, Piyush Khanna, and Omesh Tickoo. 2024.
\newblock Enhancing trust in large language models with uncertainty-aware fine-tuning.
\newblock \emph{arXiv preprint arXiv:2412.02904}.

\bibitem[{Kuhn et~al.(2023)Kuhn, Gal, and Farquhar}]{kuhn2023semantic}
Lorenz Kuhn, Yarin Gal, and Sebastian Farquhar. 2023.
\newblock \href {https://openreview.net/forum?id=VD-AYtP0dve} {Semantic uncertainty: Linguistic invariances for uncertainty estimation in natural language generation}.
\newblock In \emph{The Eleventh International Conference on Learning Representations}.

\bibitem[{Kumar et~al.(2024)Kumar, Morabito, Umbet, Kabbara, and Emami}]{kumar2024confidence}
Abhishek Kumar, Robert Morabito, Sanzhar Umbet, Jad Kabbara, and Ali Emami. 2024.
\newblock Confidence under the hood: An investigation into the confidence-probability alignment in large language models.
\newblock In \emph{Proceedings of the 62nd Annual Meeting of the Association for Computational Linguistics (Volume 1: Long Papers)}, pages 315--334.

\bibitem[{Kwiatkowski et~al.(2019)Kwiatkowski, Palomaki, Redfield, Collins, Parikh, Alberti, Epstein, Polosukhin, Devlin, Lee, Toutanova, Jones, Kelcey, Chang, Dai, Uszkoreit, Le, and Petrov}]{kwiatkowski-etal-2019-natural}
Tom Kwiatkowski, Jennimaria Palomaki, Olivia Redfield, Michael Collins, Ankur Parikh, Chris Alberti, Danielle Epstein, Illia Polosukhin, Jacob Devlin, Kenton Lee, Kristina Toutanova, Llion Jones, Matthew Kelcey, Ming-Wei Chang, Andrew~M. Dai, Jakob Uszkoreit, Quoc Le, and Slav Petrov. 2019.
\newblock \href {https://doi.org/10.1162/tacl_a_00276} {Natural questions: A benchmark for question answering research}.
\newblock \emph{Transactions of the Association for Computational Linguistics}, 7:452--466.

\bibitem[{Lambert et~al.(2025)Lambert, Morrison, Pyatkin, Huang, Ivison, Brahman, Miranda, Liu, Dziri, Lyu, Gu, Malik, Graf, Hwang, Yang, Bras, Tafjord, Wilhelm, Soldaini, Smith, Wang, Dasigi, and Hajishirzi}]{lambert2024tulu3}
Nathan Lambert, Jacob Morrison, Valentina Pyatkin, Shengyi Huang, Hamish Ivison, Faeze Brahman, Lester James~Validad Miranda, Alisa Liu, Nouha Dziri, Xinxi Lyu, Yuling Gu, Saumya Malik, Victoria Graf, Jena~D. Hwang, Jiangjiang Yang, Ronan~Le Bras, Oyvind Tafjord, Christopher Wilhelm, Luca Soldaini, Noah~A. Smith, Yizhong Wang, Pradeep Dasigi, and Hannaneh Hajishirzi. 2025.
\newblock \href {https://openreview.net/forum?id=i1uGbfHHpH} {Tulu 3: Pushing frontiers in open language model post-training}.
\newblock In \emph{Second Conference on Language Modeling}.

\bibitem[{Li et~al.(2025{\natexlab{a}})Li, Xiong, Wu, and Hooi}]{li2025conftuner}
Yibo Li, Miao Xiong, Jiaying Wu, and Bryan Hooi. 2025{\natexlab{a}}.
\newblock Conftuner: Training large language models to express their confidence verbally.
\newblock \emph{arXiv preprint arXiv:2508.18847}.

\bibitem[{Li et~al.(2025{\natexlab{b}})Li, Wang, Huang, and Liu}]{li2025graphbased}
Yukun Li, Sijia Wang, Lifu Huang, and Liping Liu. 2025{\natexlab{b}}.
\newblock \href {https://openreview.net/forum?id=BDPvuD5FTg} {Graph-based confidence calibration for large language models}.
\newblock \emph{Transactions on Machine Learning Research}.

\bibitem[{Lin et~al.(2022)Lin, Hilton, and Evans}]{linteaching}
Stephanie Lin, Jacob Hilton, and Owain Evans. 2022.
\newblock \href {https://openreview.net/forum?id=8s8K2UZGTZ} {Teaching models to express their uncertainty in words}.
\newblock \emph{Transactions on Machine Learning Research}.

\bibitem[{Liu et~al.(2020)Liu, Juanchich, Sirota, and Orbell}]{liu2020intuitive}
Dawn Liu, Marie Juanchich, Miroslav Sirota, and Sheina Orbell. 2020.
\newblock The intuitive use of contextual information in decisions made with verbal and numerical quantifiers.
\newblock \emph{Quarterly Journal of Experimental Psychology}, 73(4):481--494.

\bibitem[{Liu et~al.(2025)Liu, Yona, Caciularu, Szpektor, Rudner, and Cohan}]{liu2025metafaith}
Gabrielle Kaili-May Liu, Gal Yona, Avi Caciularu, Idan Szpektor, Tim~GJ Rudner, and Arman Cohan. 2025.
\newblock Metafaith: Faithful natural language uncertainty expression in llms.
\newblock \emph{arXiv preprint arXiv:2505.24858}.

\bibitem[{Mahaut et~al.(2024)Mahaut, Aina, Czarnowska, Hardalov, Mueller, and M{\`a}rquez}]{mahaut2024factual}
Mat{\'e}o Mahaut, Laura Aina, Paula Czarnowska, Momchil Hardalov, Thomas Mueller, and Llu{\'\i}s M{\`a}rquez. 2024.
\newblock Factual confidence of llms: on reliability and robustness of current estimators.
\newblock In \emph{Proceedings of the 62nd Annual Meeting of the Association for Computational Linguistics (Volume 1: Long Papers)}, pages 4554--4570.

\bibitem[{Mallen et~al.(2023)Mallen, Asai, Zhong, Das, Khashabi, and Hajishirzi}]{mallen-etal-2023-trust}
Alex Mallen, Akari Asai, Victor Zhong, Rajarshi Das, Daniel Khashabi, and Hannaneh Hajishirzi. 2023.
\newblock \href {https://doi.org/10.18653/v1/2023.acl-long.546} {When not to trust language models: Investigating effectiveness of parametric and non-parametric memories}.
\newblock In \emph{Proceedings of the 61st Annual Meeting of the Association for Computational Linguistics (Volume 1: Long Papers)}, pages 9802--9822, Toronto, Canada. Association for Computational Linguistics.

\bibitem[{Manakul et~al.(2023)Manakul, Liusie, and Gales}]{manakul-etal-2023-selfcheckgpt}
Potsawee Manakul, Adian Liusie, and Mark Gales. 2023.
\newblock \href {https://doi.org/10.18653/v1/2023.emnlp-main.557} {{S}elf{C}heck{GPT}: Zero-resource black-box hallucination detection for generative large language models}.
\newblock In \emph{Proceedings of the 2023 Conference on Empirical Methods in Natural Language Processing}, pages 9004--9017, Singapore. Association for Computational Linguistics.

\bibitem[{Mandel et~al.(2021)Mandel, Wallsten, and Budescu}]{mandel2021numerically}
David~R Mandel, Thomas~S Wallsten, and DV~Budescu. 2021.
\newblock Numerically bounded linguistic probability schemes are unlikely to communicate uncertainty effectively.
\newblock \emph{Earth's Future}, 9(1):e2020EF001526.

\bibitem[{Mielke et~al.(2022)Mielke, Szlam, Dinan, and Boureau}]{mielke2022reducing}
Sabrina~J Mielke, Arthur Szlam, Emily Dinan, and Y-Lan Boureau. 2022.
\newblock Reducing conversational agents’ overconfidence through linguistic calibration.
\newblock \emph{Transactions of the Association for Computational Linguistics}, 10:857--872.

\bibitem[{Milligan and Cooper(1986)}]{milligan1986study}
Glenn~W Milligan and Martha~C Cooper. 1986.
\newblock A study of the comparability of external criteria for hierarchical cluster analysis.
\newblock \emph{Multivariate behavioral research}, 21(4):441--458.

\bibitem[{Min et~al.(2020)Min, Michael, Hajishirzi, and Zettlemoyer}]{min-etal-2020-ambigqa}
Sewon Min, Julian Michael, Hannaneh Hajishirzi, and Luke Zettlemoyer. 2020.
\newblock \href {https://doi.org/10.18653/v1/2020.emnlp-main.466} {{A}mbig{QA}: Answering ambiguous open-domain questions}.
\newblock In \emph{Proceedings of the 2020 Conference on Empirical Methods in Natural Language Processing (EMNLP)}, pages 5783--5797, Online. Association for Computational Linguistics.

\bibitem[{Moxey and Sanford(2000)}]{moxey2000communicating}
Linda~M Moxey and Anthony~J Sanford. 2000.
\newblock Communicating quantities: A review of psycholinguistic evidence of how expressions determine perspectives.
\newblock \emph{Applied Cognitive Psychology: The Official Journal of the Society for Applied Research in Memory and Cognition}, 14(3):237--255.

\bibitem[{OLMo et~al.(2024)OLMo, Walsh, Soldaini, Groeneveld, Lo, Arora, Bhagia, Gu, Huang, Jordan et~al.}]{olmo20242}
Team OLMo, Pete Walsh, Luca Soldaini, Dirk Groeneveld, Kyle Lo, Shane Arora, Akshita Bhagia, Yuling Gu, Shengyi Huang, Matt Jordan, et~al. 2024.
\newblock 2 olmo 2 furious.
\newblock \emph{arXiv preprint arXiv:2501.00656}.

\bibitem[{OpenAI(2025)}]{openai2025gpt5}
OpenAI. 2025.
\newblock \href {https://openai.com/gpt-5/} {Introducing gpt-5}.
\newblock Accessed on September 8th, 2025.

\bibitem[{Schilling(2017)}]{schilling2017measures}
Ren{\'e}~L Schilling. 2017.
\newblock \emph{Measures, integrals and martingales}.
\newblock Cambridge University Press.

\bibitem[{Steyvers et~al.(2025)Steyvers, Tejeda, Kumar, Belem, Karny, Hu, Mayer, and Smyth}]{steyvers2025large}
Mark Steyvers, Heliodoro Tejeda, Aakriti Kumar, Catarina Belem, Sheer Karny, Xinyue Hu, Lukas~W Mayer, and Padhraic Smyth. 2025.
\newblock What large language models know and what people think they know.
\newblock \emph{Nature Machine Intelligence}, pages 1--11.

\bibitem[{Tang et~al.(2025)Tang, Yang, and Song}]{tangunderstanding}
Eric Tang, Bangding Yang, and Xingyou Song. 2025.
\newblock Understanding llm embeddings for regression.
\newblock \emph{Transactions on Machine Learning Research}.

\bibitem[{Tian et~al.(2024)Tian, Mitchell, Yao, Manning, and Finn}]{tian2024finetuning}
Katherine Tian, Eric Mitchell, Huaxiu Yao, Christopher~D Manning, and Chelsea Finn. 2024.
\newblock \href {https://openreview.net/forum?id=WPZ2yPag4K} {Fine-tuning language models for factuality}.
\newblock In \emph{The Twelfth International Conference on Learning Representations}.

\bibitem[{Ulmer et~al.(2024)Ulmer, Gubri, Lee, Yun, and Oh}]{ulmer-etal-2024-calibrating}
Dennis Ulmer, Martin Gubri, Hwaran Lee, Sangdoo Yun, and Seong Oh. 2024.
\newblock \href {https://doi.org/10.18653/v1/2024.acl-long.824} {Calibrating large language models using their generations only}.
\newblock In \emph{Proceedings of the 62nd Annual Meeting of the Association for Computational Linguistics (Volume 1: Long Papers)}, pages 15440--15459, Bangkok, Thailand. Association for Computational Linguistics.

\bibitem[{Ulmer et~al.(2025)Ulmer, Lorson, Titov, and Hardmeier}]{ulmer2025anthropomimetic}
Dennis Ulmer, Alexandra Lorson, Ivan Titov, and Christian Hardmeier. 2025.
\newblock Anthropomimetic uncertainty: What verbalized uncertainty in language models is missing.
\newblock \emph{arXiv preprint arXiv:2507.10587}.

\bibitem[{Vinh et~al.(2009)Vinh, Epps, and Bailey}]{vinh2009information}
Nguyen~Xuan Vinh, Julien Epps, and James Bailey. 2009.
\newblock Information theoretic measures for clusterings comparison: is a correction for chance necessary?
\newblock In \emph{Proceedings of the 26th annual international conference on machine learning}, pages 1073--1080.

\bibitem[{Vogel et~al.(2022)Vogel, Appelbaum, Haller, and Ostermann}]{vogel2022interpretation}
Hannah Vogel, Sebastian Appelbaum, Heidemarie Haller, and Thomas Ostermann. 2022.
\newblock The interpretation of verbal probabilities: A systematic literature review and meta-analysis.
\newblock \emph{German Medical Data Sciences 2022--Future Medicine: More Precise, More Integrative, More Sustainable!}, pages 9--16.

\bibitem[{Wang et~al.(2025)Wang, Friedman, Zhu, Zhu, and Mountford}]{wang2025impact}
Lifei Wang, Natalie Friedman, Chengchao Zhu, Zeshu Zhu, and S~Joy Mountford. 2025.
\newblock The impact of confidence ratings on user trust in large language models.
\newblock In \emph{Adjunct Proceedings of the 33rd ACM Conference on User Modeling, Adaptation and Personalization}, pages 365--370.

\bibitem[{Xu et~al.(2024)Xu, Wu, Diao, Liu, Wang, Chen, and Gao}]{xu2024sayself}
Tianyang Xu, Shujin Wu, Shizhe Diao, Xiaoze Liu, Xingyao Wang, Yangyi Chen, and Jing Gao. 2024.
\newblock Sayself: Teaching llms to express confidence with self-reflective rationales.
\newblock In \emph{Proceedings of the 2024 Conference on Empirical Methods in Natural Language Processing}, pages 5985--5998.

\bibitem[{Yang et~al.()Yang, Tsai, and Yamada}]{yangverbalized}
Daniel Yang, Yao-Hung~Hubert Tsai, and Makoto Yamada.
\newblock On verbalized confidence scores for llms.
\newblock In \emph{ICLR Workshop: Quantify Uncertainty and Hallucination in Foundation Models: The Next Frontier in Reliable AI}.

\bibitem[{Yona et~al.(2024{\natexlab{a}})Yona, Aharoni, and Geva}]{yona-etal-2024-large}
Gal Yona, Roee Aharoni, and Mor Geva. 2024{\natexlab{a}}.
\newblock \href {https://doi.org/10.18653/v1/2024.emnlp-main.443} {Can large language models faithfully express their intrinsic uncertainty in words?}
\newblock In \emph{Proceedings of the 2024 Conference on Empirical Methods in Natural Language Processing}, pages 7752--7764, Miami, Florida, USA. Association for Computational Linguistics.

\bibitem[{Yona et~al.(2024{\natexlab{b}})Yona, Aharoni, and Geva}]{yona2024can}
Gal Yona, Roee Aharoni, and Mor Geva. 2024{\natexlab{b}}.
\newblock Can large language models faithfully express their intrinsic uncertainty in words?
\newblock In \emph{Proceedings of the 2024 Conference on Empirical Methods in Natural Language Processing}, pages 7752--7764.

\end{thebibliography}
\bibliographystyle{acl_natbib}

\appendix
\section{Datasets and Training Details}
\label{app:hparams-data}

\subsection{Datasets}
\label{app:datasets}

We follow the preprocessing procedures described by \citet{yona-etal-2024-large} to construct our evaluation sets from PopQA and Natural Questions (NQ). We subsample a similarly sized evaluation set from TriviaQA, and we derive a complementary training set for FUT from PopQA. We share code to reproduce all datasets at \url{https://github.com/probabll/uncertainty-tuning}.%

\paragraph{PopQA.}
We preprocess PopQA\footnote{ \url{https://github.com/AlexTMallen/adaptive-retrieval/blob/main/data/popQA.tsv}.} following the same steps as \citet{yona-etal-2024-large}: we retain only \{\emph{director}, \emph{screenwriter}, \emph{producer}, \emph{author}, \emph{place of birth}, \emph{occupation}\} relations; we drop duplicate questions; and we exclude entities (subject or object) with fewer than three characters. From the resulting pool, we randomly sample 932 questions with a fixed seed (\texttt{seed=0}) to serve as our PopQA evaluation set. For training data we %
take the complement of indices (those not included in the evaluation set), yielding 5654 questions. %

\paragraph{Natural Questions (NQ).}
We again follow \citet{yona-etal-2024-large} and use the AmbigQA dataset \citep{min-etal-2020-ambigqa} in the ``light'' configuration,\footnote{\url{https://huggingface.co/datasets/sewon/ambig_qa}} which provides a non-ambiguous subset of Natural Questions. We retain only instances with type \texttt{singleAnswer}, and for each such question we take the first annotated answer string. The resulting validation split yields 932 question–answer pairs, which we use as our NQ evaluation set.

\paragraph{TriviaQA.}
We use the validation split of TriviaQA in the ``unfiltered'' configuration,\footnote{\url{https://huggingface.co/datasets/trivia_qa}} which contains questions paired with references. We shuffle with a fixed seed (\texttt{seed=42}) and retain 932 examples for evaluation. For each instance, we extract the question along with the annotated canonical answer and its aliases as references.

\subsection{Training Details}
\label{app:training-details}

\begin{table}
\centering
\begin{tabular}{ll}
\toprule
\textbf{Model} & \textbf{Learning rate} \\
\midrule
OLMo2 13B (both settings)  & $1\times 10^{-5}$ \\
OLMo2 7B (both settings)        & $5\times 10^{-6}$ \\
Tülu3 8B (both settings)        & $5\times 10^{-6}$ \\
\bottomrule
\end{tabular}
\caption{Selected learning rates for each model. FUT-interweave and FUT-postfix use the same learning rates.}
\label{tab:lrs}
\end{table}

Training data is stratified across ten confidence bins to ensure balanced supervision. Interweaving is performed using Mistral 12B\footnote{\texttt{mistralai/Mistral-Nemo-Instruct-2407}}, while postfixing uses fixed template phrases. Confidence estimation for training data construction relies on the DeBERTa Large NLI model \footnote{\texttt{microsoft/deberta-large-mnli}} \citep{he2021deberta}. Initial model generations are obtained using the \texttt{vanilla} prompt (\cref{app:generation-prompts}).

We fine-tune all models with a quantized 8-bit paged AdamW optimizer \citep{dettmers2022optimizers} and a cosine learning-rate schedule with a 10\% warmup ratio. Fine-tuning runs for three epochs. We sweep learning rates over $\{1\times10^{-4},\,1\times10^{-5},\,5\times10^{-6},\,1\times10^{-6}\}$ and select the best value based on validation loss. The selected learning rates are reported in \cref{tab:lrs}.

As our base models we use model checkpoints immediately after supervised fine-tuning. The exact Hugging Face identifiers of these checkpoints are: \texttt{allenai/OLMo-2-1124-7B-SFT}, \texttt{allenai/OLMo-2-1124-13B-SFT} and \texttt{allenai/Llama-3.1-Tulu-3-8B-SFT}.

\section{Semantic Distribution Shift}
\label{app:semantic-distribution-shift}
In \cref{sec:semantic-distribution-shift} we perform a semantic distribution shift analysis. In this Appendix, we detail the exact clustering procedure and how we evaluated its effectiveness for reproducibility purposes. 

\paragraph{Clustering algorithm.}
To group responses into semantically distinct clusters, for each response, we iterate over existing clusters and assign a response to an existing cluster if an LLM judge deems it semantically equivalent to a randomly chosen member of that cluster (prompt in \Cref{tab:sem_equivalence-prompt}). If no match is found, a new cluster is created.  %

\paragraph{Semantic distributions.}
Each response is mapped to one of the discovered semantic clusters. For a given model–prompt condition, we obtain a probability distribution over clusters by counting how many of its responses fall into each cluster and normalising by the total number of responses. The resulting distributions are defined over a shared support of clusters, though a particular model–prompt condition may assign zero probability to some clusters. We then compute the total variation distance (TVD) between these distributions to quantify semantic distribution shift for each input. Aggregating across inputs provides the results reported in Section~\ref{sec:semantic-distribution-shift}.

\begin{table}
\centering
\begin{tabular}{ll}
\toprule
\textbf{Metrics} & \textbf{Mean and St.Dev.} \\
\midrule
Rand  &  $0.960 \pm 0.039$ \\
Adjusted Rand & $0.476 \pm 0.196$ \\
MI  &  $0.924 \pm 0.068$ \\
Adjusted MI  & $0.568 \pm 0.170$ \\
\bottomrule
\end{tabular}
\caption{Evaluation of the semantic distribution shift clustering algorithm using standard clustering agreement metrics against manual annotations.}
\label{tab:clust_val}
\end{table}

\paragraph{Evaluation of clustering algorithm}
To validate the LLM-based clustering (and implicitly the LLM's performance on the task of semantic equivalence decision), we sampled 10 questions with responses from OLMo2 13B using the \texttt{vanilla} and \texttt{uncertainty} prompts. We cluster using the procedure described above, and compare those assignments against manually annotated cluster assignments. We evaluate agreement using the Rand Index and Adjusted Rand Index \citep{milligan1986study}, which measure the fraction of consistent clustering decisions, as well as Mutual Information (MI) and Adjusted MI \citep{vinh2009information}, which quantify similarity between sets of labels. Results in \Cref{tab:clust_val} show high metric consistency, supporting our procedure's reliability.

\begin{table*}
    \centering
    \begin{tabular}{p{0.95\linewidth}}
        \toprule
        \textbf{Semantic Equivalence Prompt} \\
        \midrule
\begin{lstlisting}[basicstyle=\ttfamily\tiny, breaklines=true]
I will give you a question and two candidate answers. Please determine whether the two answers 
    are semantically equivalent, or not. 
    That is, you need to check whether the two candidate answers mean the same thing, given the question. 
    Ignore any uncretainty expressions in the candidate answers; as long as the semantics of the answers are the 
    same given the question you can assume semantic equivalence. Just generate 'equivalent' or 'not equivalent'.

    Here are some examples:

    Question: What is LeBron James' profession?
    Candidate Answer 1: professional basketball player.
    Candidate Answer 2: basketball player
    Verdict: equivalent

    Question: Where was Barack Obama born?
    Candidate Answer 1: Honolulu 
    Candidate Answer 2: Hawaii
    Verdict: not equivalent

    Question: Who did Hillary Clinton marry?
    Candidate Answer 1: I doubt she married Bill Clinton.
    Candidate Answer 2: Bill Clinton.
    Verdict: equivalent

    Question: Who is the top scorer in Manchester United?
    Candidate Answer 1: David Beckham.
    Candidate Answer 2: Please use Google search for questions like this.
    Verdict: not equivalent
\end{lstlisting} \\
        \bottomrule
    \end{tabular}
    \caption{Semantic equivalence prompt used for semantic distribution shift evaluation.}
    \label{tab:sem_equivalence-prompt}
\end{table*}

\section{FUT-numerical Training Details}
\label{app:regression}
Inspired by \citet{tangunderstanding}, who exploit LLM embeddings of strings as downstream features for metric prediction, we employ a similar method to predict confidence for responses of OLMo2 13B. 
We utilise the data obtained when generating the training data of OLMo2 13B; namely questions, (unbiased) sampled answers and their confidences. 10\% of the training set is used as a validation set.
We train a multi-layer perceptron (MLP) as a regression head on top of the LLM. This model takes as input the pair $x$ - $y$ which is passed through the LLM. Then, we take the output logit layers corresponding to $y$'s tokens, collapse them into a one-dimensional vector representation (by averaging) and pass this representation to the MLP head (that has 2 hidden layers; 1024 and 512). We train using mean squared error (MSE) between the actual and predicted confidences as the loss, the AdamW optimizer, a learning rate of $1\mathrm{e}{-5}$ and a weight decay of 0.1. We train for 10 epochs and choose the epoch checkpoint with the lowest validation MSE loss. Performance is evaluated using the cMFG metric \citep{yona-etal-2024-large}, replacing decisiveness with predicted confidence values, with results reported in Table~\ref{tab:regression} of the main paper.

\begin{figure}
  \centering
  \includegraphics[width=\linewidth]{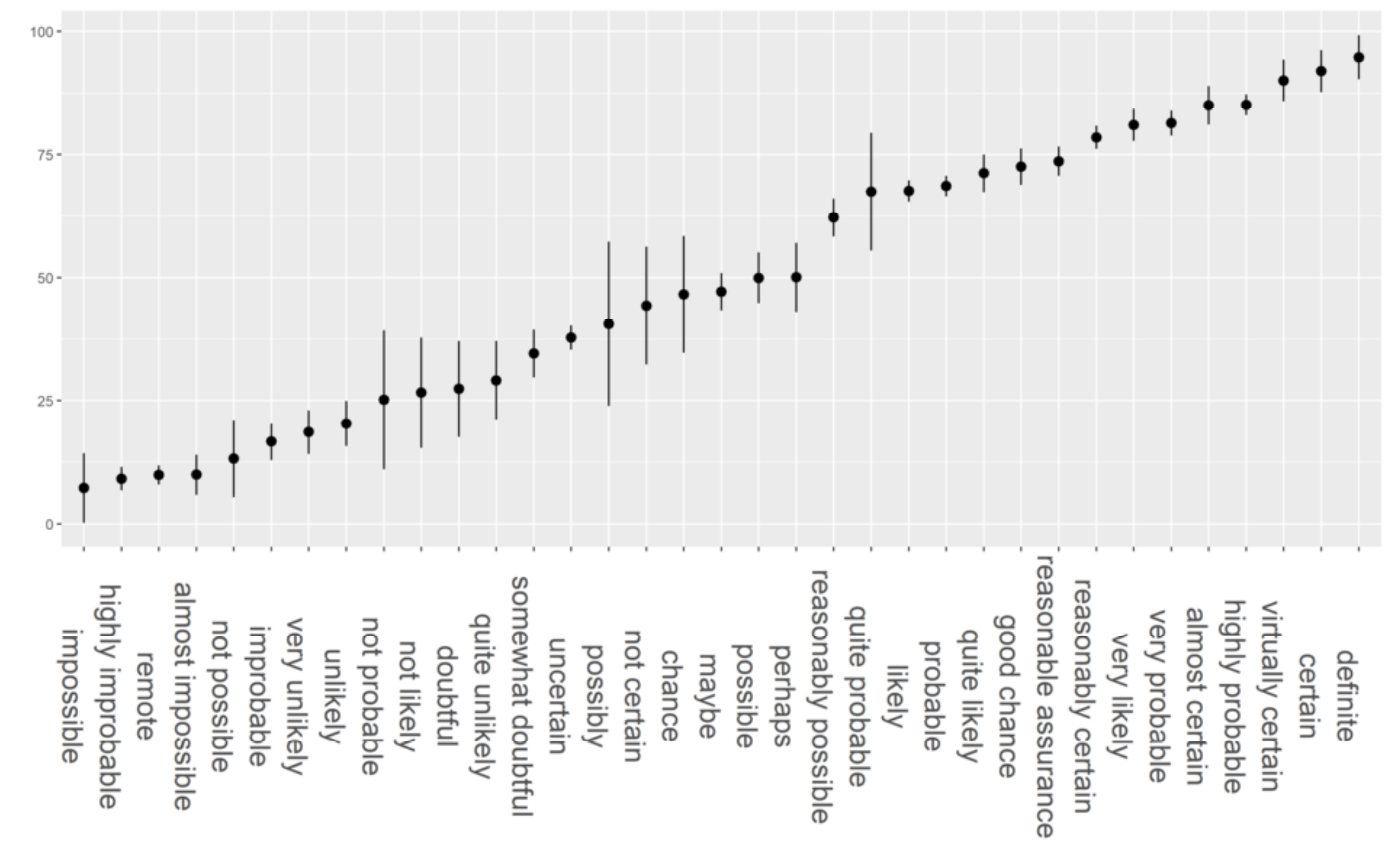}
  \caption{An exact reproduction of Figure~5 from \citet{vogel2022interpretation}: averaged mean values (with 95\% confidence intervals) for 35 verbal probability expressions aggregated across prior studies.}
  \label{fig:vogel-fig5}
\end{figure}

\section{Verbal-Numerical Probability Correspondences}
\label{app:vogel}
To ground our linguistic hedges in established human interpretations, we adopt the verbal-numerical probability correspondences summarised in the meta-analysis by \citet{vogel2022interpretation}. Table~\ref{tab:conf_to_hedger} in the main paper lists the specific bins we use for mapping model confidence to hedge phrases (and postfix variants). For reference, \Cref{fig:vogel-fig5} reproduces the original summary figure from \cite[Figure~5]{vogel2022interpretation}, showing averaged mean values (and 95\% confidence intervals) for 35 commonly used expressions arranged approximately linearly from ``impossible'' to ``definite''. This mapping underpins both our interweave and postfix strategies and aligns the decisiveness scale used by the judge with human interpretations reported in the literature.

\section{Prompts}
\label{app:prompts}

\subsection{LLM Judge Prompts}
\label{app:judge-prompts}
In computing faithfulness we use LLMs to decompose responses into assertions, extract decisiveness from model generations, and perform NLI inferences for confidence estimation. 
Our prompts follow the general setup of \citet{yona-etal-2024-large}, but with two notable differences. 
First, when evaluating a response $y$, we decompose it into individual assertions $a_{1}, \ldots, a_{K}$. We then compare each assertion $a_{i}$ against the full text of $N$ model samples $s_{1}, \ldots, s_{N}$. In contrast to \citet{yona-etal-2024-large}, who also decompose the samples into assertions, we operate at the full-sample level. If any statement in a sample $s_{j}$ contradicts an assertion $a_{i}$, we count this as a contradiction of that assertion.
This simplifies the pipeline without noticeably affecting reliability, since the vast majority of samples ($\geq 98\%$) contain only a single assertion.
Second, for NLI we adapt the prompt to explicitly remove uncertainty hedges (\textit{e.g.} \textit{likely}, \textit{impossible}) before comparison. 
Without this step, NLI models can misinterpret uncertainty hedges like ``almost impossible'' as contradictions, even when the underlying assertions agree. 
The exact prompts used are provided in \cref{tab:judge-prompts}.

\subsection{Generation Prompts}
\label{app:generation-prompts}
For our experiments we rely on two simple brevity-inducing prompts, adapted from \citet{yona-etal-2024-large}. 
The \texttt{vanilla} prompt enforces brevity and is used as a baseline in evaluation, as the template for training data construction and when generating from fine-tuned FUT models. 
The \texttt{uncertainty} prompt adds an explicit instruction to hedge when uncertain, serving as a stronger prompting baseline at evaluation time. 
While inspired by the prompts of \citet{yona-etal-2024-large}, our versions differ slightly in wording. %
For completeness and reproducibility, we outline them below:
\begin{small}
\begin{itemize}
    \item \textbf{Vanilla:}
    \texttt{Answer the following question with a brief, one sentence long answer.}
    
    \texttt{Question: <question>}
    
    \texttt{Answer:}
    
    \item \textbf{Uncertainty:}
    \texttt{Answer the following question using a succinct 
    (at most one sentence) and full answer.
    If you are uncertain about your answer to the question, 
    convey this uncertainty linguistically by precisely 
    hedging this answer.}

    \texttt{Question: <question>}
    
    \texttt{Answer:}
\end{itemize}
\end{small}

\begin{table*}[t]
\centering
\begin{tabular}{p{0.48\linewidth} p{0.48\linewidth}}
\toprule
\textbf{Decisiveness Extraction Prompt} & \textbf{NLI Prompt} \\
\midrule
\begin{lstlisting}[basicstyle=\ttfamily\tiny, breaklines=true]
You will be given a question and a proposed answer. Your task is to extract any assertions made by the proposed answer; for each assertion, you should output a decisiveness score: a number in [0,1] representing how likely you think this assertion is to be true, judged solely based on the provided answer. You should pay special attention to the usage of any hedging modifiers in the original answer, used to convey uncertainty in the truthfulness of the assertion. If the proposed answer punts the question, the extracted assertion should be the empty string and the decisiveness score should be 1.0.

Here are some examples:

Question: criminal minds episode where JJ becomes a profiler
Proposed answer: It's unlikely that JJ officially becomes a profiler in Criminal Minds season 6, episode 3.
Extracted assertion: JJ officially becomes a profiler in Criminal Minds season 6, episode 3.
Decisiveness score: 0.3.

Question: in criminal minds, does JJ becomes a profiler
Proposed answer: Yes, JJ becomes a profiler during the series.
Extracted assertion: JJ becomes a profiler during the series.
Decisiveness score: 1.0.

Question: in criminal minds, is JJ a profiler for the entire series?
Proposed answer: I am certain that JJ is not a profiler for the entire series. I am reasonably certain that JJ officially becomes a profiler in Criminal Minds season 6.
Extracted assertion: JJ is not a profiler for the entire series.
Decisiveness score: 1.0.
Extracted assertion: JJ officially becomes a profiler in Criminal Minds season 6.
Decisiveness score: 0.8.

Question: criminal minds episode where JJ becomes a profiler
Proposed answer: I am somewhat doubtful that the episode in which JJ officially becomes a profiler in Criminal Minds is episode 2 in season 6.
Extracted assertion: JJ officially becomes a profiler in Criminal Minds season 6, episode 2.
Decisiveness score: 0.4.

Question: In criminal minds, does JJ becomes a forensic psychologist?
Proposed answer: It's almost impossible that JJ becomes a forensic psychologist.
Extracted assertion: JJ becomes a forensic psychologist
Decisiveness score: 0.2.

Question: <question>
Proposed answer: <response>
\end{lstlisting}
&
\begin{lstlisting}[basicstyle=\ttfamily\tiny, breaklines=true]
I will give you a question and two candidate answers. Please determine whether the two answers are contradicting. If one of the Candidate Answers is a canned response (as in, avoids answering the question), then the verdict should be "no contradiction". Answers may contain epistemic markers to hedge their responses, e.g. "It's likely", "It's impossible", "It's somewhat doubtful", "It is certain", ignore these in your verdict: compare the assertions made stripped from their uncertainty hedgers. Before making your verdict, strip the answers of uncertainty hedgers and produce Stripped Answer 1 and Stripped Answer 2. Pay extra attention to phrases such as "It is impossible": this is an uncertainty hedger indicating low confidence in the output, ignore the hedger and base your NLI verdict only on statement following it. When stripping answers of uncertainty hedgers, do not add negations. "It's highly improbably that A" should map to "A", *not* to "not A".

Here are some examples:

Question: Varicose veins occur when what happens to the veins under the skin?
Candidate Answer 1: becomes enlarged.
Candidate Answer 2: becomes irritated by something.
Stripped Answer 1: becomes enlarged.
Stripped Answer 2: becomes irritated by something.
Verdict: contradiction

Question: What is LeBron James' profession?
Candidate Answer 1: It's likely that he is a professional basketball player.
Candidate Answer 2: It's very unlikely that he is a basketball player.
Stripped Answer 1: He is a professional basketball player.
Stripped Answer 2: He is a basketball player.
Verdict: no contradiction

Question: Where was Barack Obama born?
Candidate Answer 1: Honolulu
Candidate Answer 2: Hawaii
Stripped Answer 1: Honolulu
Stripped Answer 2: Hawaii
Verdict: no contradiction

Question: Who did Hillary Clinton marry?
Candidate Answer 1: It's likely that she married Bill.
Candidate Answer 2: It is certainly Bill Clinton.
Stripped Answer 1: She married Bill.
Stripped Answer 2: Bill Clinton.
Verdict: no contradiction

Question: What position does David Beckham typically play?
Candidate Answer 1: It's impossible that he was a right winger.
Candidate Answer 2: It's somewhat doubtful that he was a striker.
Stripped Answer 1: He was a right winger.
Stripped Answer 2: He was a striker.
Verdict: contradiction

Question: Who is the top scorer in Manchester United?
Candidate Answer 1: David Beckham.
Candidate Answer 2: Please use Google search for questions like this.
Stripped Answer 1: David Beckham.
Stripped Answer 2: Please use Google search for questions like this.
Verdict: no contradiction

Question: How many movies did Brad Pit star in?
Candidate Answer 1: Likely over 80 movies.
Candidate Answer 2: Probably 75
Stripped Answer 1: Over 80 movies.
Stripped Answer 2: 75
Verdict: contradiction

Question: <question>
Candidate Answer 1: <answer-1>
Candidate Answer 2: <answer-2>
Produce Stripped Answer 1, Stripped Answer 2 and Verdict.
\end{lstlisting}
\\
\bottomrule
\end{tabular}
\caption{Prompts used in our evaluation. The left column shows the \textit{decisiveness extraction prompt}, adapted from \citet{yona-etal-2024-large} and aligned with the verbal–numerical mappings of \citet{vogel2022interpretation}. The right column shows the \textit{NLI prompt}, also adapted from \citet{yona-etal-2024-large} to strip uncertainty hedges from model generations to avoid spurious contradictions.}
\label{tab:judge-prompts}
\end{table*}

\section{Belief Preservation\label{app:belief-preservation}}

\emph{Context.} The function $f$  (\ie, steps 1--5 of \cref{sec:faithful-uncertainty-tuning}) is constructed to preserve the `semantics' or asserted contents of its argument, modulo any hedging. This is realised by step 4 of the procedure (true in FUT-postfix, contingent on the quality of the LLM approach to interweaving in FUT-interweave). %
We can express this explicitly with the help of a semantic map $n(\cdot)$, which analyses its argument in an abstract semantic space capturing the assertions regardless of surface form and hedging (\wrt \cref{fig:sketch2}, $n(\cdot)$ identifies a standard name, such as \emph{Elmo}, \emph{Grover} and \emph{Oscar}, for the entity blamed as responsible for pushing \emph{Big Bird}). Then, step 4 of $f$ implies that $n(y) = n(f(y))$.
The distribution over asserted contents (\eg,  probabilities over clusters in \cref{fig:sketch2}) is the distribution we get when we map responses from $p$ through $n$; it too corresponds to a pushforward, namely, $p_{\#n}$.
\emph{Claim.} The pushforward $p_{\#f}$ preserves $p$'s beliefs of asserted contents, that is, pushing $p_{\#f}$ by $n$ identifies the same distribution as $p_{\#n}$. 
\emph{Proof sketch.} %
For $Y\sim p(\cdot|x)$ and $R\sim p_{\#f}(\cdot|x)$, and, by the pushforward's conservation of total mass \citep{schilling2017measures}, it follows that $n(Y) \sim n(R)$, or, equivalently, $p_{\#n} \sim p_{\#f\#n}$. In words: in the semantic space onto which $n(\cdot)$ maps responses, the distributions of samples obtained via $Y\sim p(\cdot|x)$ or $R\sim p_{\#f}(\cdot|x)$ are equivalent.

\end{document}